\documentclass{article} 
\usepackage[nonatbib,preprint]{main_sty}

\usepackage{amsmath,amsfonts,bm}

\def\eqref#1{equation~\ref{#1}}

\def\1{\bm{1}}

\DeclareMathAlphabet{\mathsfit}{\encodingdefault}{\sfdefault}{m}{sl}
\SetMathAlphabet{\mathsfit}{bold}{\encodingdefault}{\sfdefault}{bx}{n}

\usepackage{natbib}
\usepackage{hyperref}
\usepackage{url}

\usepackage{microtype}
\usepackage{graphicx}
\usepackage{booktabs} 

\usepackage{hyperref}

\usepackage{amsmath}
\usepackage{amssymb}
\usepackage{mathtools}
\usepackage{amsthm}

\usepackage{listings}
\usepackage[version=4]{mhchem}

\usepackage{cleveref}
\crefname{equation}{Eq.}{Eqs.}
\crefname{table}{Table}{Tables}
\crefname{figure}{Figure}{Figures}
\crefname{section}{Section}{Sections}
\crefname{algorithm}{Algorithm}{Algorithms}

\theoremstyle{plain}

\theoremstyle{definition}

\theoremstyle{remark}

\usepackage{xcolor,colortbl}
\usepackage{adjustbox}
\usepackage{url}
\usepackage{multirow,multicol,xspace}
\usepackage{makecell}
\usepackage{float}
\usepackage{graphics}
\usepackage{paralist}
\usepackage{wrapfig}
\usepackage[normalem]{ulem}
\usepackage{pgfplots}
\usepackage{subcaption}
\pgfplotsset{width=8cm,compat=1.17} 
\usepackage{pifont}

\usepackage{arydshln}
\makeatletter
\def\adl@drawiv#1#2#3{%
        \hskip.5\tabcolsep
        \xleaders#3{#2.5\@tempdimb #1{1}#2.5\@tempdimb}%
                #2\z@ plus1fil minus1fil\relax
        \hskip.5\tabcolsep}
\newcommand{\cdashlinelr}[1]{%
  \noalign{\vskip\aboverulesep
           \global\let\@dashdrawstore\adl@draw
           \global\let\adl@draw\adl@drawiv}
  \cdashline{#1}
  \noalign{\global\let\adl@draw\@dashdrawstore
           \vskip\belowrulesep}}
\makeatother

\definecolor{gray}{gray}{0.92}
\definecolor{LightCyan}{rgb}{0.88,1,1}

\title{Multimodal Large Language Models for Inverse Molecular Design with Retrosynthetic Planning}

\author{%
  Gang Liu$^{1,4}$, \quad Michael Sun$^{2,4}$, \quad Wojciech Matusik$^2$, \quad  Meng Jiang$^1$, \quad Jie Chen$^3$ \\
  $^1$University of Notre Dame \quad $^2$MIT CSAIL \quad $^3$ MIT-IBM Watson AI Lab \\
  $^4$ This work was done while GL and MS interned at \\ the MIT-IBM Watson AI Lab, IBM Research \\
  \texttt{\{gliu7, mjiang2\}@nd.edu}, \\
  \texttt{\{msun415, wojciech\}@csail.mit.edu}, \\
  \texttt{chenjie@us.ibm.com} \\ 
}

\newcommand{\method}{Llamole\xspace}

\begin{document}

\maketitle

\begin{abstract}
While large language models (LLMs) have integrated images, adapting them to graphs remains challenging, limiting their applications in materials and drug design. This difficulty stems from the need for coherent autoregressive generation across texts and graphs. To address this, we introduce Llamole, the first multimodal LLM capable of interleaved text and graph generation, enabling molecular inverse design with retrosynthetic planning.
Llamole integrates a base LLM with the Graph Diffusion Transformer and Graph Neural Networks for multi-conditional molecular generation and reaction inference within texts, while the LLM, with enhanced molecular understanding, flexibly controls activation among the different graph modules. 
Additionally, Llamole integrates A* search with LLM-based cost functions for efficient retrosynthetic planning. We create benchmarking datasets and conduct extensive experiments to evaluate Llamole against in-context learning and supervised fine-tuning. Llamole significantly outperforms 14 adapted LLMs across 12 metrics for controllable molecular design and retrosynthetic planning. 

\end{abstract}


\section{Introduction}
\label{sec:introduction}
\begin{wrapfigure}{r}{0.35\textwidth}
  \centering
  \vspace{-0.25in}
  \includegraphics[width=0.35\textwidth]{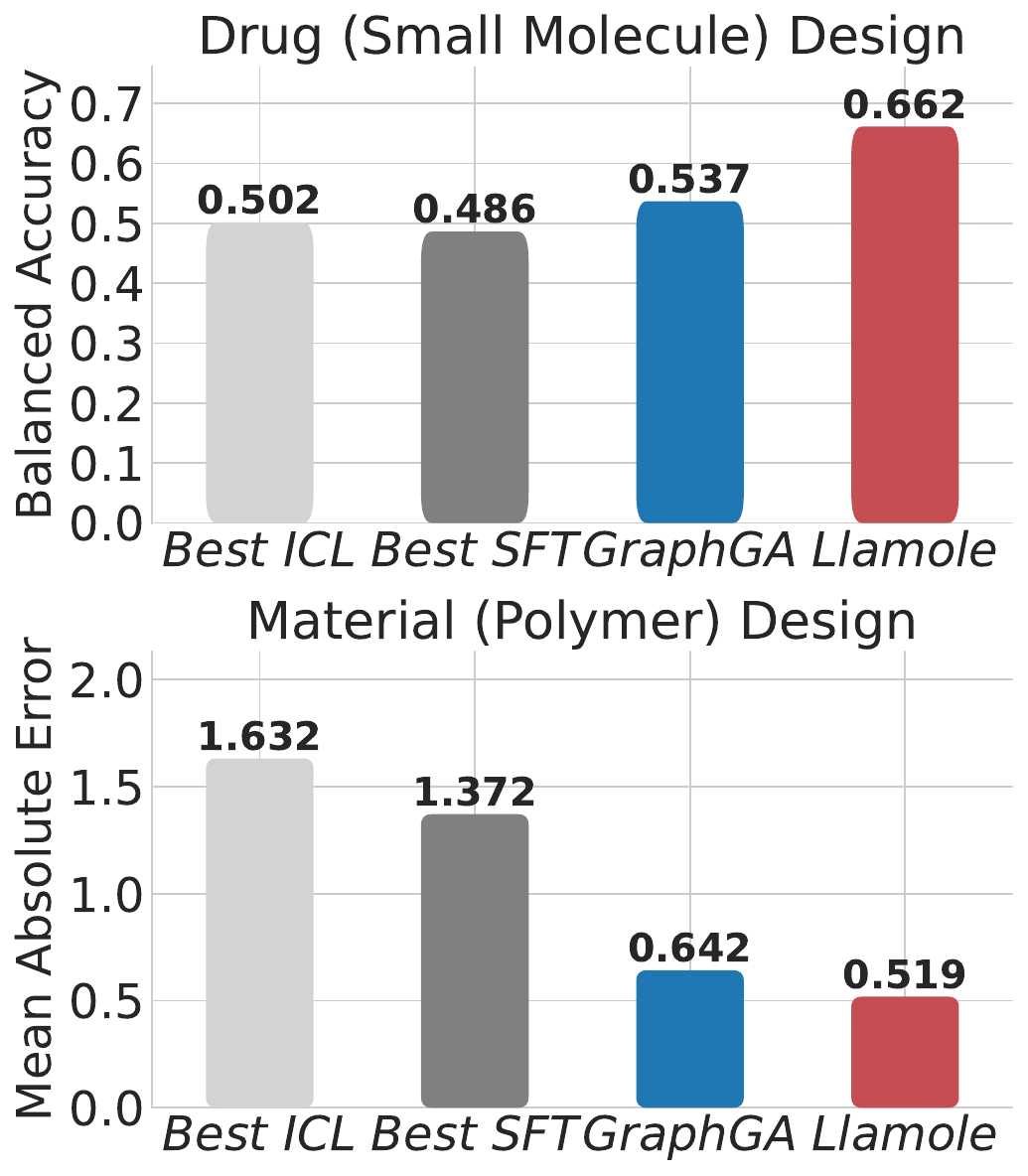}
  \vspace{-0.3in}
  \caption{Comparison of Controllability: Results are averaged from the best numbers from~\cref{tab:design-performance}.}
  \label{fig:compare_graphga}
  \vspace{-0.15in}
\end{wrapfigure}

Large language models (LLMs) are transforming machine learning across many domains~\citep{achiam2023gpt}. Their success in natural language processing has extended to areas handling not only text but also image and speech data~\citep{dong2023dreamllm,wu2024nextgpt}, thanks to the development of multimodal capabilities. Recently, the potential of LLMs for molecular discovery has been actively explored~\citep{jablonka202314}. However, LLMs struggle in the chemical domain, exhibiting poor generation quality and planning capability~\citep{guo2023can}. This is due to the unique, graph structures of molecular data, which are challenging for LLMs that typically handle sequential texts.

\begin{figure}[!t]
    \centering
    \vspace{-0.1in}
    \includegraphics[width=0.9\textwidth]{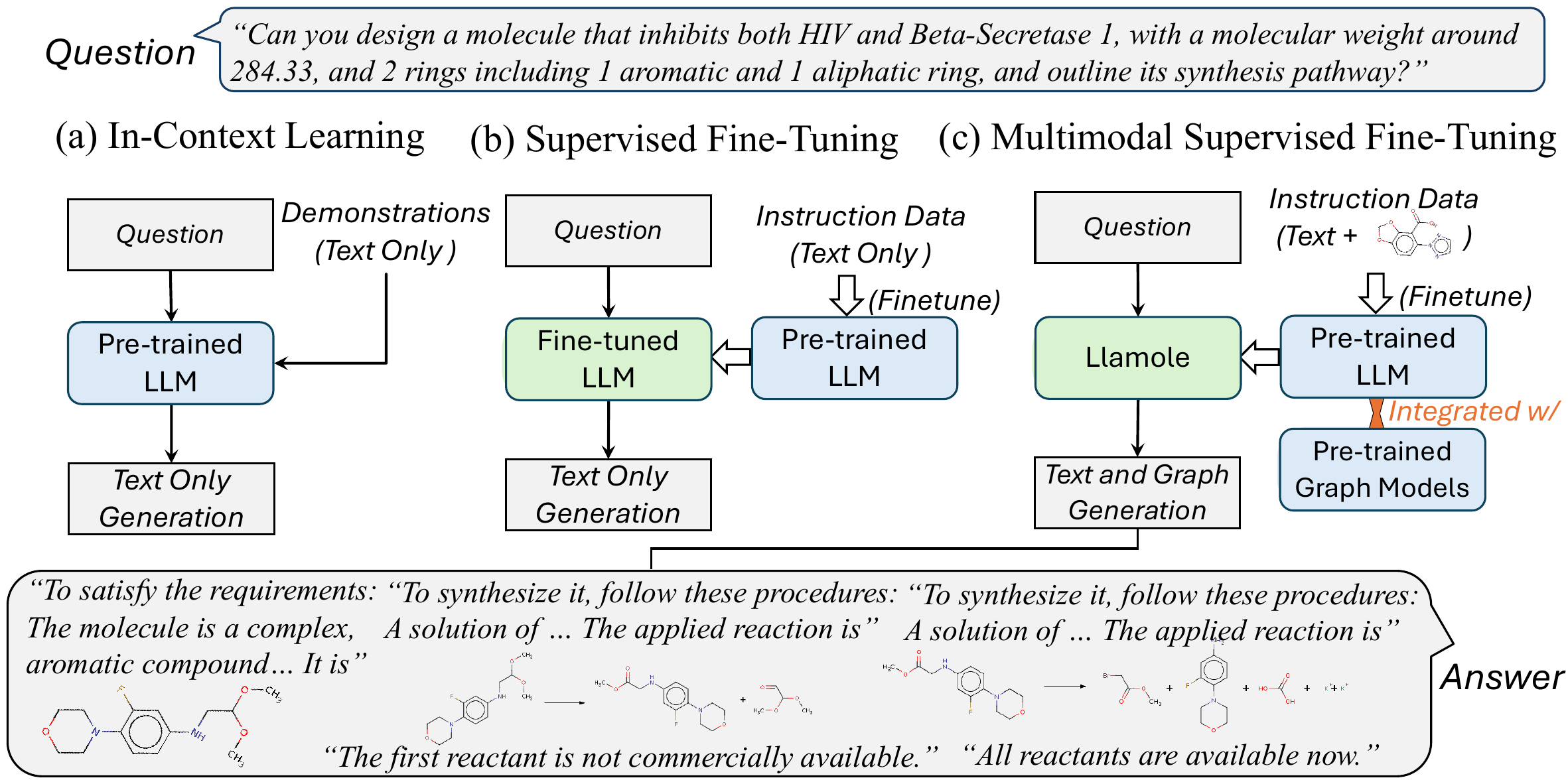}
    \caption{Three LLM-based methods for molecular design. The question outlines requirements for properties, structures, and synthesis, addressed as follows: (a) In-Context Learning and (b) Supervised Fine-Tuning use text-only data for demonstrations and instruction tuning, respectively. (c) The proposed \method uses graph-text multimodal data to fine-tune the LLM, integrating parameter-frozen graph models for interleaved text and graph generation with reaction inference.}
    \label{fig:idea}
    \vspace{-0.2in}
\end{figure}

Inverse molecular design requires LLMs to be controllable for generating molecular structures that meet multi-property and synthesizability requirements~\citep{chen2020retro,gao2021amortized}. These requirements can be detailed as questions for LLM input, as shown in~\cref{fig:idea}. 
Answering these questions demands a comprehensive understanding of molecular structures and their relationship to properties. However, sequence-based LLMs struggle with this because they are pre-trained or fine-tuned solely on text representations of molecules, e.g., SMILES~\citep{weininger1988smiles}. 
To illustrate this, we investigate 14 LLMs for molecular generation in~\cref{fig:compare_graphga} across 10K drug and material questions: ten using in-context learning (ICL) and four with supervised fine-tuning (SFT). LLMs generate molecular structures based on the questions, and their properties are obtained through oracles to assess the differences between requirements and generated outputs. Details of the experimental set-ups and results can be found in \cref{sec:experiment}. In summary, even the best LLMs perform worse than GraphGA~\citep{gao2022sample}, a simple yet effective graph-based method, in designing molecules with satisfactory properties.

As illustrated in~\cref{fig:idea}, practical answers for molecular design are more complex than what can be achieved by using graph methods or LLMs alone. The generation begins with a paragraph describing the intended molecule for multi-conditional generation, followed by retrosynthetic planning, detailing each synthesis step—one reaction per paragraph—in reverse order, from the target molecule to purchasable reactants. Thus, multimodal LLMs (MLLMs) are essential, with LLMs handling text generation and graph models managing molecular design.

In this work, we propose the multimodal \textbf{L}arge \textbf{la}nguage model for \textbf{mole}cular discovery (\textbf{\method}). As shown in~\cref{fig:idea} (c), the model seamlessly integrates LLMs and graph models within a multimodal autoregressive framework, enabling the interleaved generation of text, molecules, and reactions. It predicts the next token across both word and chemical spaces, framed as multi-class prediction tasks for word vocabulary, atom/bond types, and reaction templates. For retrosynthetic planning, \method integrates A* search to efficiently identify synthesis pathways for the designed molecule.

To implement \method, we augment a base LLM with two pre-trained graph modules: the Graph Diffusion Transformer (Graph DiT) for multi-conditional molecule generation~\citep{liu2024inverse} and a GNN for reaction template prediction. The base LLM controls the generation flow using a trigger-query-prediction approach with two sets of trigger tokens for the Graph DiT and GNN, respectively. Upon predicting a trigger token, one or a few query tokens summarize the prior text as vectors, activating the corresponding graph modules and generating molecules or predicting reaction templates. Afterward, the base LLM can resume text generation, aided by a graph encoder that encodes the previously generated molecule. In retrosynthetic planning, the LLM computes heuristics to efficiently assist the A* search in navigating the vast reaction space for multi-step generation.

Our work has several highlights. First, \method is the first MLLM for molecular design, capable of interleaved generation of text and graphs. Second, we curated a dataset along with fine-tuning instructions to benchmark complex yet realistic molecular design outcomes, including human conversation. Third, we present compelling experimental results that demonstrate the competitiveness of \method against 14 LLMs and GraphGA, as shown in \cref{fig:compare_graphga}. With details in~\cref{tab:design-performance,tab:retro-success-ratio}, \method improves LLM performance by up to 80.9\% across 12 metrics for controllable molecular generation and increases the success rate for retrosynthetic planning from 5.5\% to 35\%.

\section{Preliminaries}
\subsection{Autoregressive Language Modeling}
Given a sequence of word tokens $W = \{w_1, w_2, \dots, w_L \}$ of length $L$ from the vocabulary $\mathcal{W}$, LLMs parameterized by $\theta_1$ decompose the joint distribution as $p_{\theta_1}({W}) = \prod_{i=1}^{L} p_{\theta_1}(w_i|{W}_{<i})$, where $W_{<i}$ represents the tokens preceding the $i$-th position.
These models are optimized by minimizing the negative log-likelihood between their predictions and the empirical data distribution, resulting in:
\begin{equation}\label{loss-lm-next-token}
    \mathcal{L}_\text{LM} = \sum_i -\log p_{\theta_1} (w_i|{W}_{<i}).
\end{equation}

\subsection{Molecular Design with Graph Diffusion Models}
Molecular graphs can be modeled through diffusion in discrete spaces~\citep{austin2021structured,vignac2022digress,liu2024inverse}. Given a one-hot encoded data point $\mathbf{x} \in \mathbb{R}^F$ with $F$ categories (e.g., a node or an edge), discrete models perform diffusion using a transition matrix $\mathbf{Q}$, where $[\mathbf{Q}^t]_{ij} = q(\mathbf{x}^t_j \mid \mathbf{x}^{t-1}_i)$ for $i,j \in [1,F]$.
The forward diffusion with $\mathbf{Q}$ is: $q(\mathbf{x}^{t} \mid \mathbf{x}^{t-1}) = \operatorname{Cat}(\mathbf{x}_t; \mathbf{p}=\mathbf{x}^{t-1}\mathbf{Q}^{t})$, where $\operatorname{Cat}(\mathbf{x}; \mathbf{p})$ denotes the categorical distribution over $\mathbf{x}$ with probabilities given by $\mathbf{p}$.
Starting from the original data point $\mathbf{x}=\mathbf{x}^0$, we have $q(\mathbf{x}^t \mid \mathbf{x}^0) = \operatorname{Cat}\left(\mathbf{x}^t; \mathbf{p}=\mathbf{x}^0 \bar{\mathbf{Q}}^t\right)$, where $\bar{\mathbf{Q}}^t = \prod_{i\leq t} \mathbf{Q}^{i}$. The forward diffusion gradually corrupts data points. When the total timestep $T$ is large enough, $q(\mathbf{x}^T)$ converges to a stationary distribution.
The reverse process samples from $q(\mathbf{x}^T)$ and gradually removes noise. The posterior distribution $q(\mathbf{x}^{t-1} \mid \mathbf{x}^{t})$ is calculated as $q(\mathbf{x}^{t-1}|\mathbf{x}^t, \mathbf{x}^0) \propto \mathbf{x}^t (\mathbf{Q}^t)^\top \odot \mathbf{x}^0 \bar{\mathbf{Q}}^{t-1}$. Using a denoising model parameterized by $\theta_2$, this posterior can be approximated by $p_{\theta_2}(\mathbf{x}^{t-1}|\mathbf{x}^t, \mathbf{x}^0)$. For inverse molecular design with multi-property constraints, the denoising model can be optimized by minimizing the negative log-likelihood for $\mathbf{x}^0$:
\begin{equation}\label{loss-discrete-diffusion} \mathcal{L}_\text{DM} = \mathbb{E}_{q(\mathbf{x}^0)} \mathbb{E}_{q(\mathbf{x}^t \mid \mathbf{x}^0)} \left[-\log p_{\theta_2} \left( \mathbf{x}^0 \mid c_1, c_2, \dots, c_M, \mathbf{c}_\text{text}, \mathbf{x}^t \right) \right],
\end{equation}
where $M$ molecular properties are denoted by $\{c_i\}_{i=1}^M$, and the text embedding is $\mathbf{c}_\text{text}$. These conditions can be handled by Graph DiT~\citep{liu2024inverse} without introducing additional predictors for guidance~\citep{ho2022classifier}.

\subsection{One-Step Reaction Prediction with Graph Neural Networks}
Retrosynthesis needs to predict the reverse of a synthetic reaction, which decomposes chemical products into reactants. A GNN parameterized by $\theta_3$ takes the product $G_\text{product}$ to predict the label $r \in \mathcal{R}$ in the reaction space $\mathcal{R}$. This label is interpreted as the template and determines the reactants. With the text condition $\mathbf{c}_\text{text}$, we minimize the negative log-likelihood of the label distribution $q(r)$:
\begin{equation}\label{loss-reaction-label-predict} 
\mathcal{L}_\text{predictor} = \mathbb{E}_{q(r)} \left[- \log p_{\theta_3} (r \mid \mathbf{c}_\text{text}, G_\text{product})\right]. 
\end{equation}

\subsection{Retrosynthetic Planning with A* Search}
Given molecules from the structure space $\mathcal{G}$, a subset $\mathcal{G}_\text{avail}$ represents available molecular structures that can be purchased as building blocks for synthesis. For any target $G_\text{target}$, one-step prediction of the reversed reaction may not yield reactants within $\mathcal{G}_\text{avail}$. Thus, retrosynthesis typically requires multi-step planning to find pathways from building blocks to the target in reverse order. The search space of chemical reactions can be navigated using A* on an AND-OR tree $\mathcal{T}$, with $G_\text{target}$ as the root. Reaction nodes follow an ``AND'' relation, requiring all child reactants, while molecule nodes follow an ``OR'' relation, meaning the product can be synthesized by any child reaction~\citep{chen2020retro}.

\textbf{Selection:} We select nodes from the frontier $\mathcal{F}(\mathcal{T})$ containing unexplored molecule nodes to expand the tree. Given an oracle cost function $J(\cdot)$, the next node is selected as $G_\text{next} = \arg\min_{G \in \mathcal{F}(\mathcal{T})} J(G)$ to minimize the cost. A well-designed $J(\cdot)$ improves search efficiency and aids in global optimality.

\textbf{Expansion:} After selecting $G_\text{next}$, a single GNN predictor call can generate many one-step retrosynthesis proposals. The GNN provides top-candidate reaction templates, each linked to different reactants. Thus we can form molecule nodes under the reaction node as an AND-OR stump.

\textbf{Update and Cost:} After expanding $G_\text{next}$, the tree becomes $\mathcal{T}^\prime$. We update the nodes in $\mathcal{T}^\prime$ for the next iteration. A* selects the path that minimizes $J(\cdot)=J_\text{current}(\cdot) + J_\text{heuristic}(\cdot)$, which includes the cost from the start to the current node $J_\text{current}(\cdot)$ and a heuristic estimate of the cost to the goal $J_\text{heuristic}(\cdot)$. With the GNN predictor, the negative log-likelihood of the reaction can be used to compute path cost $J_\text{current}(\cdot)$ to the leaf molecule node, we design $J_\text{heuristic}(\cdot)$ with the LLM in \method.




\section{Llamole: Multimodal Large Language Model for Molecular Discovery}
\begin{figure}[t]
    \centering
    \includegraphics[width=0.9\textwidth]{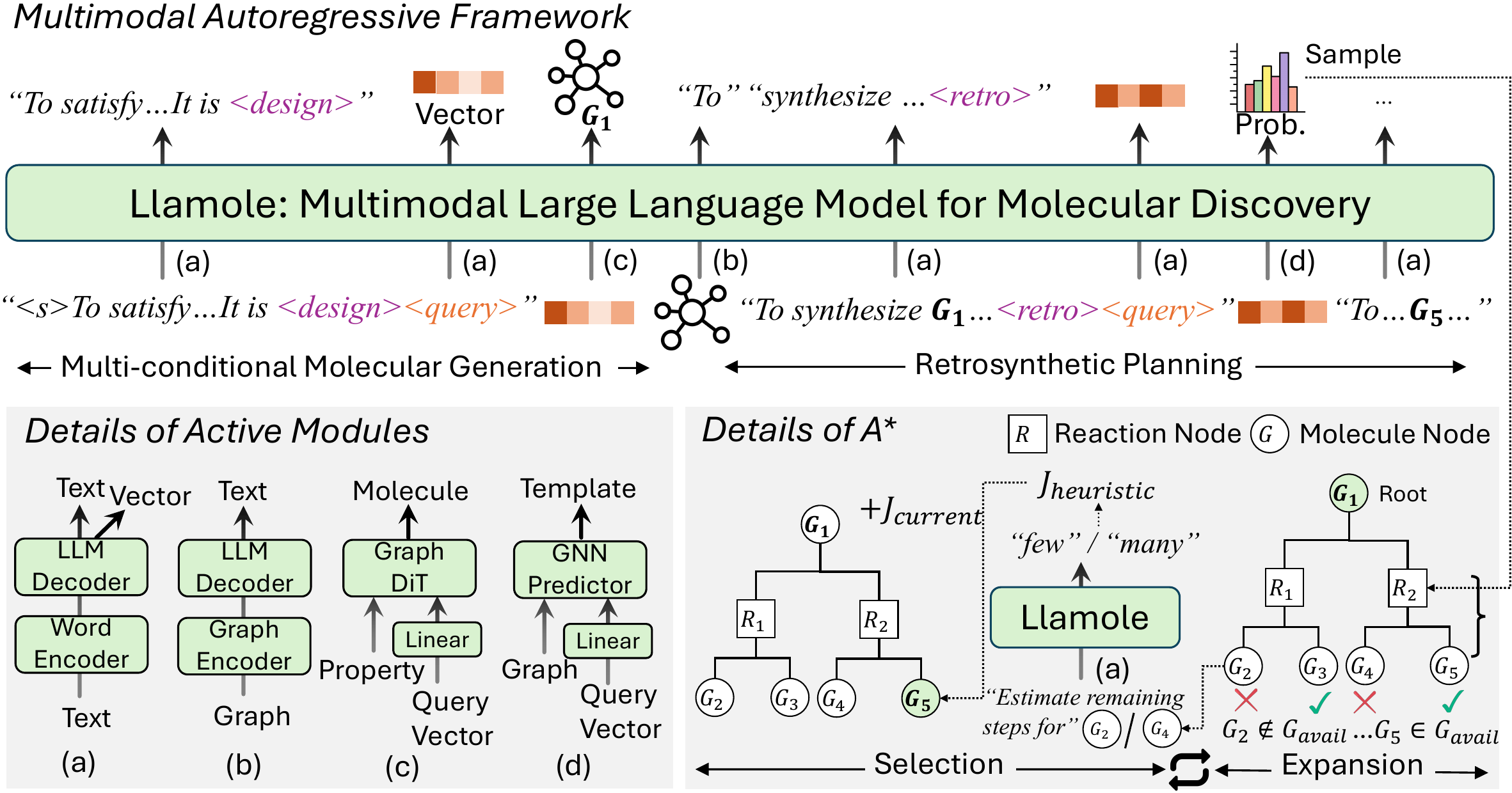}
    \vspace{-0.1in}
    \caption{
    Overview of \method: Trigger tokens (\texttt{<design>} and \texttt{<retro>}) switch active modules from the base LLM to the respective graph component. The subsequent \texttt{<query>} token utilizes output vectors from the LLM to summarize past texts as conditions. Using these, \method generates molecules and predicts one-step reactions. Enhanced with a graph encoder and A* search, \method efficiently plans synthesis routes through selection and expansion iterations on the AND-OR Tree.}
    \label{fig:model}
    \vspace{-0.1in}
\end{figure}

\subsection{Multimodal Autoregressive Modeling}
In molecular discovery, the sequence may include molecular structures $\mathcal{G}$ and retrosynthetic reactions $\mathcal{R}$
with each molecule or reaction tokenized. The sequence $Y = \{y_1, y_2, \dots, y_N\}$, where $y_i \in \mathcal{W} \cup \mathcal{G} \cup \mathcal{R}$, combines these tokens. The sequence is interleaved with tokens in different spaces. 
Suppose the molecule appears at position $i$; then, we typically see:  
$$
\ldots, \quad Y_i \in \mathcal{G}, \quad Y_{i+1:i+L} \in \mathcal{W}, \quad Y_{i+L+1} \in \mathcal{R}, \quad \ldots
$$
where $L$ is the length of the text following the molecule at position $i$. The sequence starts with text. If position $i$ denotes the first molecule in the sequence, then $Y_{<i} \in \mathcal{W}$; otherwise, $y_{i-1} \in \mathcal{R}$.
To handle non-word tokens, we integrate domain-specific Graph DiT and GNN with the LLM, forming a multimodal LLM, i.e., \method. Parameterized by $\Theta$, \method unifies the cross-entropy losses from~\cref{loss-lm-next-token,loss-discrete-diffusion,loss-reaction-label-predict} into autoregressive modeling:
\begin{equation}\label{loss-mllm}
    \mathcal{L}_\text{\method} = \mathcal{L}_\text{LM} + \mathcal{L}_\text{DM} + \mathcal{L}_\text{predictor} = \sum_i
    - \log p_\Theta (y_i|Y_{<i}).
\end{equation}
$\mathcal{L}_\text{DM}$ interprets ${Y}_{<i}$ as the input conditions, including desirable molecular properties and text conditions $\{c_i\}_{i=1}^M \cup \{\mathbf{c}_\text{text}\}$ for the autoregression of $Y_i$ in $\mathcal{G}$.
In $\mathcal{L}_\text{predictor}$, ${Y}_{<i}$ represents $G_\text{product}$ and $\mathbf{c}_\text{text}$. 
Here, $G_\text{product}$ is generated from previous diffusion models or as intermediate $G \notin \mathcal{G}_\text{avail}$ in retrosynthesis. The autoregression for the label $Y_i$ is performed in the reaction space $\mathcal{R}$.

We present an overview of multimodal autoregression with \method in~\cref{fig:model}, divided into controllable molecular generation and retrosynthetic planning. The base LLMs perform multiple roles: generating text, controlling the switch of active modules, and providing cost functions for A* search. Augmented with the graph models, the overall parameters in \method are $\Theta = \{\theta_1, \theta_2, \theta_3, \phi_1, \phi_2, \phi_3\}$, where $\phi_1$ and $\phi_2$ project text into $\mathbf{c}_\text{text}$ for the Graph DiT and GNN predictor, respectively. The graph encoder with $\phi_3$ projects molecule tokens into the LLM.
Next, we detail the design space of \method.


\subsection{\method Design Space}
\method consists of a base LLM and two pre-trained graph modules: the Graph DiT for molecule generation and the GNN for one-step reaction prediction. The base LLM employs a trigger-query-prediction approach using two sets of special tokens to switch between modules.

\textbf{Trigger Tokens.} 
\method defines two special trigger tokens to augment the word vocabulary $\mathcal{W}$: \texttt{<design>} for switching between the LLM and Graph DiT, and \texttt{<retro>} for switching between the LLM and GNN predictor. When a trigger token is predicted, \method activates the corresponding graph model. After molecule generation or reaction prediction, the active modules revert to the LLM.

\textbf{Query Tokens.} 
We introduce another set of special tokens, named query tokens \texttt{<query>} automatically placed after triggers. They use the LLM to query previous tokens and output hidden states as $\mathbf{c}_\text{hidden}$. A linear layer is applied: $\mathbf{c}_\text{text} = \operatorname{Linear}(\mathbf{c}_\text{hidden})$, adjusting the input size for the graph models. We use different query tokens for different triggers. Query tokens allow us to share parameters $\phi_1$ and $\phi_2$ with $\theta_1$, enhancing both efficiency and effectiveness. We can apply ensemble methods by repeating the query tokens multiple times and averaging the $\mathbf{c}_\text{hidden}$ values~\citep{dong2023dreamllm}.

Besides the special tokens, \method enhances molecule understanding with a graph encoder and uses the LLM to provide the cost function in A* search for retrosynthetic planning.
 
\textbf{Graph Encoder.} 
The graph encoder parameterized by $\phi_3$ replaces the word encoder in the LLM tokenizer for molecule tokens. The LLM decoder takes molecule embeddings from the graph encoder, along with text embeddings from the tokenizer, into the Transformer layers for next token generation. We use a pre-trained Graph Isomorphism Network (GIN)~\citep{xu2018powerful} as the graph encoder, optimized via molecule-text contrastive learning similar to CLIP~\citep{radford2021learning}.

\textbf{A* Cost Function with LLM.}
We define $J_\text{heuristic}$ as a multi-choice problem, where each choice, assigned a score, represents synthesis complexity, from few to many steps. The LLM estimates the remaining synthesis steps for the leaf molecule node $G \in \mathcal{F}(\mathcal{T}) \setminus \mathcal{G}_\text{avail}$ in the search tree $\mathcal{T}$. It outputs probabilities for each choice, and $J_\text{heuristic}$ is computed as the weighted score by averaging the scores with their probabilities. For $G \in \mathcal{F}(\mathcal{T}) \cap \mathcal{G}_\text{avail}$, $J_\text{heuristic}=0$.


\subsection{End-to-End Model Fine-Tuning and Generation}

\textbf{Supervised Fine-Tuning.} 
We use multimodal SFT to connect the base LLM and other graph modules in \method~\citep{ouyang2022training}. Specifically, we freeze the parameters for the graph modules ($\theta_2$ and $\theta_3$) and fine-tune the LLM parameters $\theta_1$, the learnable special tokens, and the linear layers for the query tokens ($\phi_1$ and $\phi_2$). We freeze the parameters of the pre-trained graph encoder ($\phi_3$) and add a tunable linear layer between it and the LLM decoder. The optimization can be conducted end-to-end with~\cref{loss-mllm}. The SFT aligns the LLM with domain-specific graph models. To maintain generality in the base LLM, we employ parameter-efficient LoRA~\citep{hu2021lora}.

\textbf{Interleaved Generation.}
Given a question as shown in~\cref{fig:idea}, \method performs controllable and synthesizable molecular designs, as presented in~\cref{fig:model}. For the controllable generation, \method uses the base LLM to analyze the requirements and switches to the Graph DiT for generating $G_\text{target}$ when the trigger is predicted. For the synthesizable generation, \method plans synthesis routes for $G_\text{target}$.
A* search on the AND-OR tree $\mathcal{T}$ aids in multi-step generation, interleaving molecule and reaction nodes, with $G_\text{target}$ as the root. During each selection-expansion iteration, A* selects $G_\text{next} = \arg\min_{G \in \mathcal{F}(\mathcal{T})} J(G)$ from the leaf nodes $\mathcal{F}(\mathcal{T})$. The graph encoder embeds molecule tokens into the LLM, which generates reaction conditions until the token \texttt{<retro>} is triggered, activating the GNN predictor. The predictor then predicts the top-50 templates as reaction nodes, along with corresponding reactants as molecule nodes for the next iteration.
A* stops after finding a route from $G_\text{target}$ to $\mathcal{G}_\text{avail}$ with satisfying all AND-OR constraints, or if it fails after 30 seconds or 300 iterations. Upon success, the text with the corresponding reaction along the route is returned for retrosynthesis; otherwise, the base LLM directly generates texts.

\section{Benchmarking for Multimodal Molecular Design}\label{sec:setup-data-model}
\begin{figure}[t]
    \centering    \includegraphics[width=0.86\textwidth]{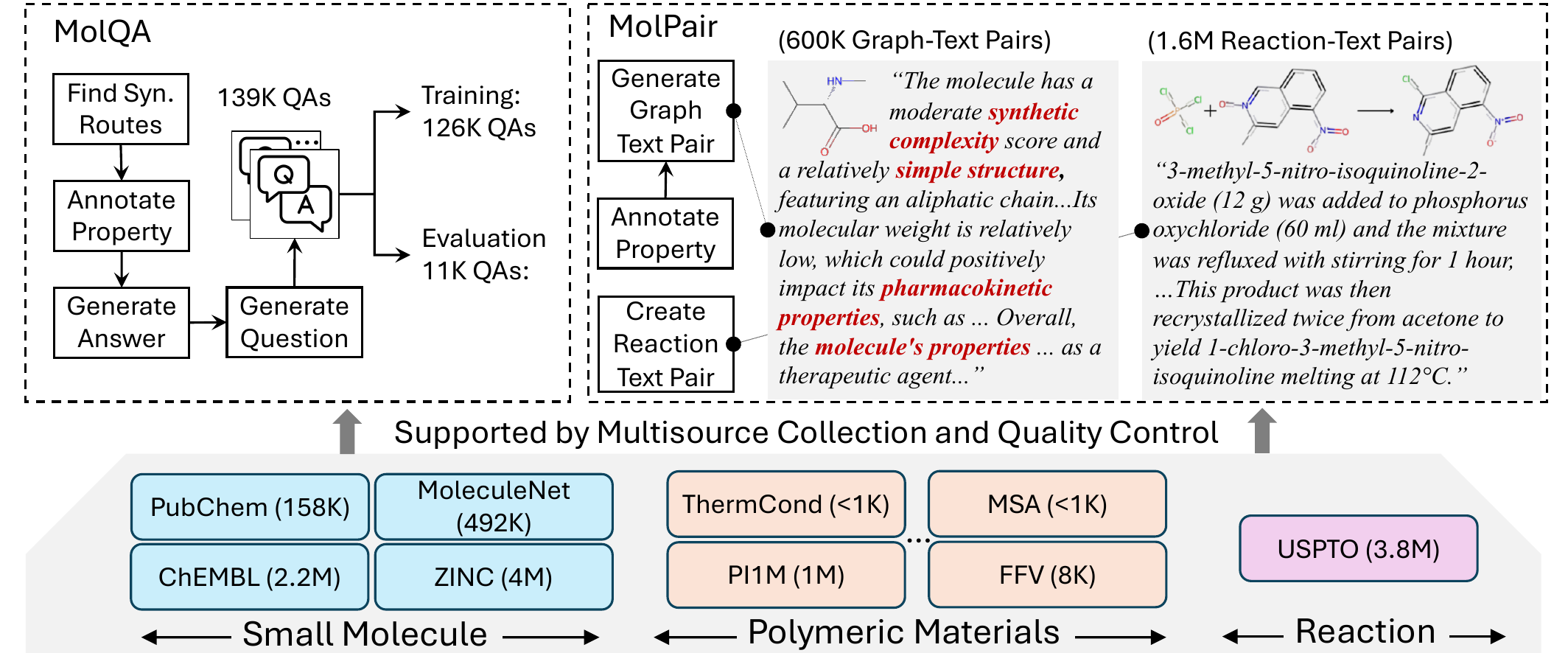}
    \caption{Creation of MolQA and MolPair: MolQA comprises two sets: a training set for ICL and (multimodal) SFT, and a test set for evaluation. MolPair consists of graph-text and reaction-text pairs, with red highlights indicating synthetic complexity, structure, and properties information.}
    \label{fig:benchmark}
    \vspace{-0.25in}
\end{figure}

To train Llamole, we need instruction data that provide detailed language supervision and evaluation covering synthetic complexity, drug and material utility, and reaction conditions. However, existing data based on PubChem~\citep{kim2021pubchem} are only usable for small molecules and lack such details. Thus, we create MolQA, a large-scale graph-text multimodal instruction dataset for systematic LLM benchmarking used in~\cref{sec:experiment}. We also create MolPair with graph-text and reaction-text pairwise data to pre-train graph modules, as detailed in~\cref{sec:add-pretrain}. To this end, we first collect multisource molecule data (\cref{fig:benchmark}), with details in~\cref{sec:add-benchmark}. Then we create MolQA and MolPair.

\textbf{MolQA: Instruction Data Creation.} 
USPTO reactions include text descriptions. We use Enamine's 1.3 million small molecules as $\mathcal{G}_\text{avail}$. The depth-first search identifies routes from reaction products (i.e. target molecules) to molecules within $\mathcal{G}_\text{avail}$, resulting in about 139K routes with lengths ranging from 1 to 10. We sample around 11K routes (750 for materials and 9986 for drugs) for testing and use the rest for instruction tuning. We focus on eight popular properties for benchmarking (i.e, $M=8$ for $\{c_i\}_1^M$ in~\cref{loss-discrete-diffusion}). They include three drug-related categorical properties~\citep{wu2018moleculenet}: (1) HIV virus replication inhibition (HIV), (2) blood-brain barrier permeability (BBBP), and (3) human $\beta$-secretase 1 inhibition (BACE) and five continuous material properties~\citep{thornton2012polymer}: (4) CO$_2$ permeability, (5) N$_2$ permeability, (6) O$_2$ permeability, (7) fractional free volume (FFV), and (8) thermal conductivity (TC).
Not all target molecules have these properties. To enrich properties and texts, two supervised GNNs predict drug and material properties with confidence scores as in~\citet {liu2022graph,liu2023semi}. Only high-confident predictions are selected for annotation. Llama-3-70B then generates descriptions using a template that incorporates these properties with structural and synthesis information from toolkits like RDKit.
There are no polymerization reactions; we consider the monomer structure of the polymer as the synthesis target. 
We assemble molecule descriptions, text, and reactions from synthesis routes as answer data. Then Llama-3-70B is prompted to generate questions, resulting in MolQA with the example as shown in~\cref{fig:idea}. Details are in~\cref{sec:add-create-molqa}.

\textbf{MolPair: Pairwise Data Creation.}
After excluding the target molecules from the instruction data, we use the remaining text-reaction data from USPTO to pre-train the GNN reaction predictor. Similarly, we utilize all other small molecules and polymers to pre-train the Graph DiT and graph encoder. For generalization, we expand beyond the eight properties used in the instruction data. For drug utility, we train another GNN to predict 41 properties, including toxicity, safety, enzyme interaction, absorption, distribution, metabolism, excretion (ADME), and biological activity~\citep{swanson2024admet}. For material utility, we consider 14 properties, such as thermal, physical, thermodynamic, permeability, solubility, and dielectric properties. Llama-3-70B generates related texts for these properties, incorporating structural and synthetic information. Finally, there are around 600K graph-text pairs for both small molecules and polymers to support pre-training. Details are in~\cref{sec:add-create-molpair}.



\section{Experiment}\label{sec:experiment}
We conduct a systematic evaluation to demonstrate \method's superior performance in controllable and synthesizable molecular design (RQ1). We investigate \method's performance in controllable molecular generation through ablation and case studies (RQ2). We analyze retrosynthetic performance of LLMs, focusing on error analysis and the efficiency and effectiveness of \method (RQ3).



\textbf{Set-ups:}
We include LLM baselines from 7B to 70B, such as Llama, Mistral, Qwen, Granite, and Flan-T5, using either ICL or LoRA-based SFT. The MolQA test set contains 9,986 QA pairs for material design and 750 for drug design. LLMs are prompted with questions to generate responses for texts, molecules, and reactions. For controllability, we evaluate up to 12 metrics across four aspects: (1) chemical validity, (2) similarity to the reference based on Morgan fingerprints~\citep{rogers2010extended}, (3) BLEU-4 and ROUGE-L scores against reference texts, and (4) deviation from desired properties. We follow~\cite{gao2022sample} to use well-trained random forests as the oracle functions for obtaining properties of designed molecules. We focus on three drug-related categorical properties assessed by balanced accuracy (BA) and five continuous material properties assessed by mean absolute error (MAE). For retrosynthesis, we evaluate the success rate of designed molecules against those available in \(\mathcal{G}_\text{avail}\) from Enamine. Details are in~\cref{sec:add-exp-setup}.




\begin{table*}[t!]
\renewcommand{\arraystretch}{1.2}
\renewcommand{\tabcolsep}{1mm}
\caption{Multi-Conditional Molecular Design with LLMs: Best overall results in each metric are in \colorbox{red!15}{\textbf{bold}}, best baseline results are in \colorbox{gray}{\textit{italic}}. Balanced Accuracy (BA) $=\frac{\text{True Positive Rate} + \text{True Negative Rate}}{2}$.}
\centering
\begin{adjustbox}{width=0.98\textwidth}
\begin{tabular}{lcccccccccccc}
\toprule
\multirow{2}{*}{\shortstack{Base LLM \\ or Method}} & \multicolumn{2}{c}{Structure ($\uparrow$)} & \multicolumn{2}{c}{Text ($\uparrow$)} & \multicolumn{3}{c}{Drug (BA $\uparrow$)} & \multicolumn{5}{c}{Material (MAE $\downarrow$)} \\
\cmidrule(lr){2-3} \cmidrule(lr){4-5} \cmidrule(lr){6-8} \cmidrule(lr){9-13}
& Validity & Similarity & BLEU-4 & ROUGE-L & HIV & BBBP & BACE & CO$_2$Perm & N$_2$Perm & O$_2$Perm & FFV & TC \\
\midrule
GraphGA & \cellcolor{gray}\textit{0.885} & 0.112 & NA & NA & \cellcolor{gray}\textit{0.536} & \cellcolor{gray}\textit{0.515} & 0.560 & \cellcolor{gray}\textit{0.847} & \cellcolor{gray}\textit{1.556} & \cellcolor{gray}\textit{0.747} & \cellcolor{red!15}\textbf{0.020} & \cellcolor{gray}\textit{0.042} \\
\midrule
\multicolumn{13}{l}{\cellcolor{gray!20}\textbf{In-Context Learning}} \\
Llama-2-7B & 0.167 & 0.024 & 0.030 & 0.141 & 0.051 & 0.060 & 0.053 & 5.463 & 3.982 & 4.943 & 0.308 & 0.199 \\
Mistral-7B & 0.251 & 0.044 & 0.066 & 0.203 & 0.163 & 0.153 & 0.200 & 5.062 & 3.824 & 4.657 & 0.289 & 0.186 \\
Qwen2-7B & 0.180 & 0.012 & 0.030 & 0.147 & 0.089 & 0.091 & 0.085 & 5.552 & 4.251 & 5.068 & 0.322 & 0.211 \\
Llama-3-8B & 0.656 & 0.112 & 0.155 & 0.307 & 0.471 & 0.473 & \cellcolor{gray}\textit{0.562} & 3.233 & 3.106 & 2.924 & 0.171 & 0.123 \\
Flan-T5-XXL & 0.570 & 0.094 & \cellcolor{gray}\textit{0.226} & \cellcolor{gray}\textit{0.388} & 0.329 & 0.333 & 0.403 & 2.869 & 3.039 & 2.799 & 0.165 & 0.120 \\
Granite-13B & 0.498 & 0.079 & 0.170 & 0.326 & 0.260 & 0.293 & 0.285 & 2.994 & 3.165 & 2.993 & 0.180 & 0.123 \\
Llama-2-13B & 0.346 & 0.058 & 0.121 & 0.279 & 0.236 & 0.250 & 0.259 & 5.031 & 4.285 & 4.816 & 0.291 & 0.184 \\
Mistral-8x7B & 0.546 & 0.094 & 0.181 & 0.345 & 0.345 & 0.346 & 0.388 & 3.695 & 3.150 & 3.440 & 0.191 & 0.138 \\
Llama-2-70B & 0.299 & 0.045 & 0.099 & 0.222 & 0.237 & 0.242 & 0.274 & 5.368 & 4.336 & 5.017 & 0.319 & 0.202 \\
Llama-3-70B & 0.706 & 0.124 & 0.210 & 0.367 & 0.415 & 0.403 & 0.484 & 2.659 & 2.848 & 2.421 & 0.135 & 0.099 \\
\midrule
\multicolumn{13}{l}{\cellcolor{gray!20}\textbf{Supervised Fine-tuning}} \\
Mistral-7B & 0.718 & 0.125 & 0.105 & 0.216 & 0.460 & 0.483 & 0.515 & 3.269 & 3.094 & 2.985 & 0.184 & 0.128 \\
Qwen2-7B & 0.768 & 0.133 & 0.221 & 0.377 & 0.436 & 0.457 & 0.457 & 2.691 & 2.562 & 2.721 & 0.147 & 0.106 \\
Llama-3-8B & 0.797 & \cellcolor{gray}\textit{0.136} & 0.093 & 0.206 & 0.426 & 0.445 & 0.440 & 2.222 & 2.322 & 2.119 & 0.110 & 0.086 \\
Llama-3.1-8B & 0.692 & 0.121 & 0.121 & 0.250 & 0.417 & 0.432 & 0.433 & 3.210 & 2.991 & 2.974 & 0.179 & 0.122 \\
\midrule
\multicolumn{13}{l}{\cellcolor{gray!20}\textbf{Llamole}} \\
Mistral-7B & 0.900 & 0.139 & \cellcolor{red!15}\textbf{0.262} & \cellcolor{red!15}\textbf{0.434} & 0.596 & 0.617 & 0.740 & \cellcolor{red!15}\textbf{0.593} & 1.409 & 0.565 & 0.021 & 0.028 \\
Qwen2-7B & 0.888 & 0.135 & 0.261 & 0.432 & 0.600 & \cellcolor{red!15}\textbf{0.639} & \cellcolor{red!15}\textbf{0.746} & 0.645 & 1.452 & 0.581 & 0.021 & \cellcolor{red!15}\textbf{0.026} \\
Llama-3.1-8B & \cellcolor{red!15}\textbf{0.913} & \cellcolor{red!15}\textbf{0.142} & 0.254 & 0.427 & \cellcolor{red!15}\textbf{0.623} & 0.629 & 0.713 & 0.653 & \cellcolor{red!15}\textbf{1.344} & \cellcolor{red!15}\textbf{0.549} & 0.021 & 0.030 \\
\midrule
\multicolumn{13}{l}{\cellcolor{gray!20}\textbf{Improvement of Llamole (\%)}}  \\
vs. All & +3.2 & +4.4 & +15.9 & +11.9 & +16.2 & +24.1 & +32.7 & +22.9 & +6.7 & +22.2 & -5.0 & +28.6 \\
vs. LLMs & +14.6 & +4.4 & +15.9 & +11.9 & +32.3 & +32.3 & +32.7 & +70.6 & +37.5 & +72.6 & +80.9 & +65.1 \\
\bottomrule
\end{tabular}
\end{adjustbox}
\label{tab:design-performance}
\end{table*}
\begin{figure}[t]
  \centering
  \vspace{-0.2in}
  \begin{subfigure}{0.49\textwidth}
    \centering
    \includegraphics[width=\textwidth]{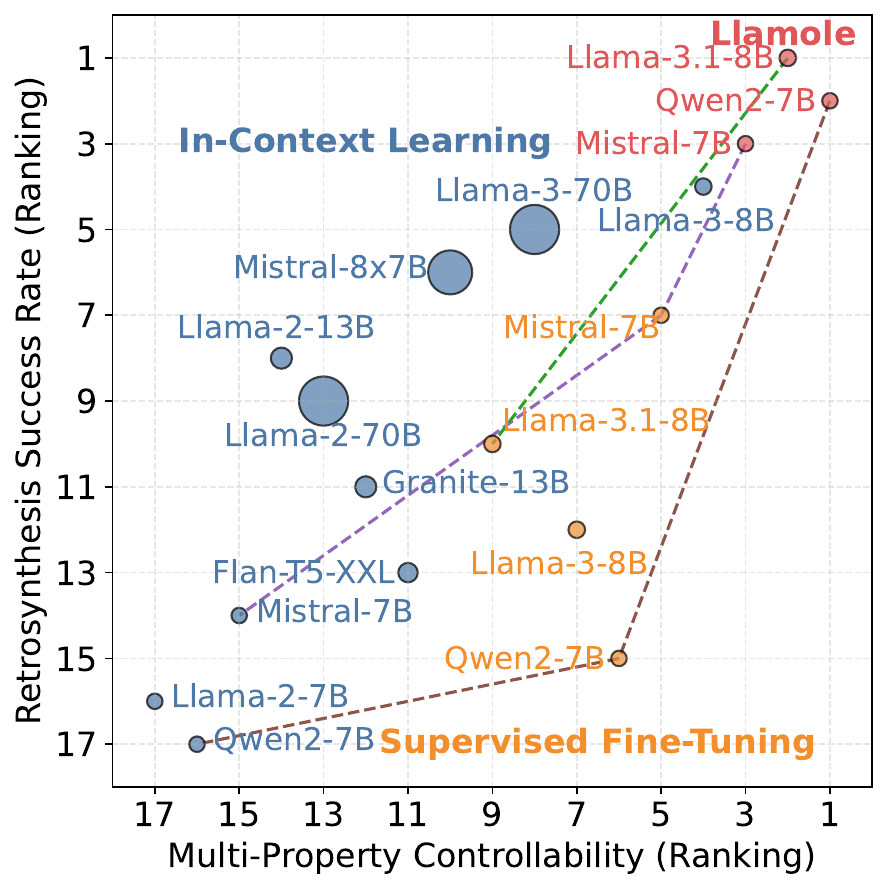}
    \vspace{-0.2in}
    \caption{LLM for Drug (Small Molecule) Design}
    \label{fig:llm_drug}
  \end{subfigure}
  \hfill
  \begin{subfigure}{0.49\textwidth}
    \centering
    \includegraphics[width=\textwidth]{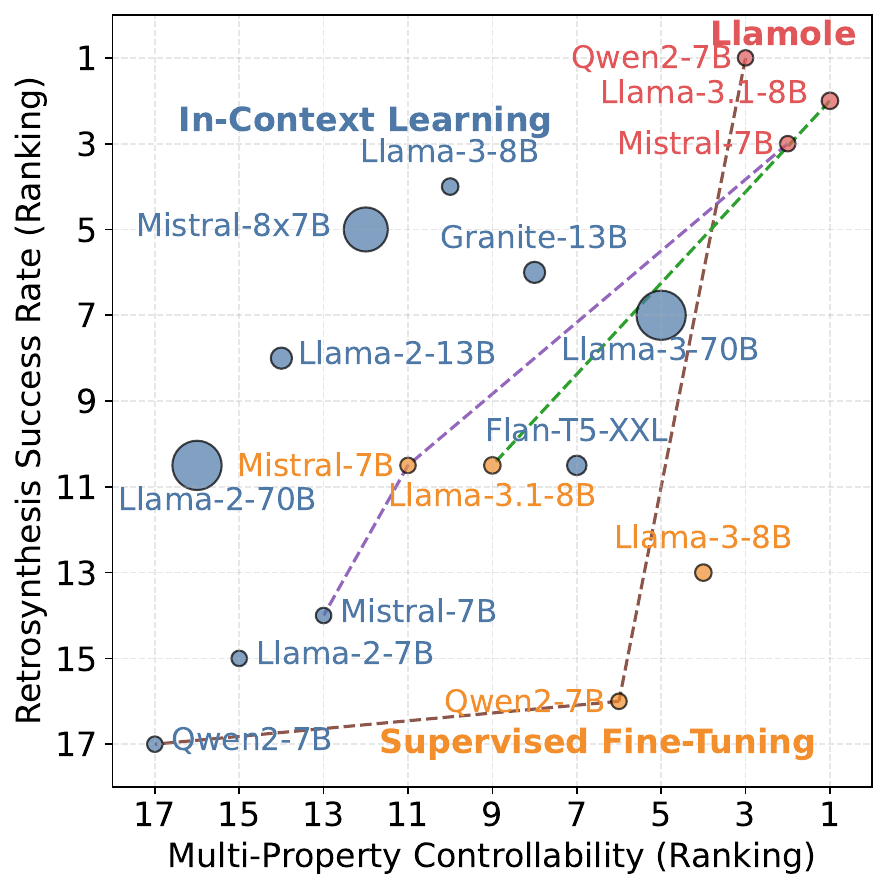}
    \vspace{-0.2in}
    \caption{LLM for Material (Polymer) Design}
    \label{fig:llm_material}
  \end{subfigure}
  \caption{Overall Comparison of LLMs for Controllability and Synthesizability: Performance is ranked by averaged BA/MAE (x-axis) and retrosynthesis success rate (y-axis). Circle size indicates model size. LLMs with ICL, SFT, and \method are highlighted in blue, orange, and red, respectively.}
  \label{fig:rank_perfomance}
\end{figure}

\subsection{RQ1: LLMs for Controllable and Synthesizable Molecular Design}
\label{sec:exp-rq1-discovery}

\cref{tab:design-performance} and~\cref{tab:retro-success-ratio} detail LLM performance in controllability and retrosynthesis success rate. The overall performance rankings are summarized in~\cref{fig:rank_perfomance}. Our key observations are:

(1) \textbf{\method significantly outperforms other LLMs in text generation, controllable molecule generation, and retrosynthetic planning.} \method fine-tuned on various 7B-parameter LLMs, as shown in~\cref{tab:retro-success-ratio}, results in top-3 rankings, surpassing 70B models that are 10$\times$ larger across all 12 metrics for controllability and planning success. Specifically, \method enhances chemical structure validity by 14.6\%, structure controllability by 4.4\%, and text generation by 11.9\%-15.9\%. Additionally, \method improves property controllability by 32\% to 80\%. In retrosynthesis, \cref{tab:retro-success-ratio} indicates \method increases the success ratio from 5\% to 35\% for drugs and to 17.9\% for polymers.

(2) \textbf{SFT improves molecular design but may not always enhance retrosynthesis.} According to~\cref{fig:rank_perfomance,tab:design-performance}, SFT enables 7B LLMs to achieve chemical validity, structure, and property control comparable to 70B LLMs with ICL. However, it offers minimal improvement in planning ability for the generated target molecule. A notable example is Llama-3-8B from~\cref{tab:retro-success-ratio}, where SFT reduces its retrosynthesis planning success from 5.5\% to below 1\%. Except for Llama-3-8B, we connect LLM performance with the same baseline but different learning methods in~\cref{fig:rank_perfomance}. The results show that SFT methods still outperform ICL with the same base 7B models in most cases.

\begin{table}[t]
\renewcommand{\arraystretch}{1}
\renewcommand{\tabcolsep}{1mm}
\caption{Retrosynthetic Success Rate: Best results are in \colorbox{red!15}{\textbf{bold}}, best baseline results are in \colorbox{gray}{\textit{italic}}.}
\label{tab:retro-success-ratio}
\centering
\resizebox{\columnwidth}{!}{%
\begin{tabular}{@{}c@{}}
\toprule
\begin{tabular}{@{}lccccccccc@{}}
& \multicolumn{9}{c}{\textbf{In-Context Learning}} \\
\cmidrule(l){2-10}
& Llama-2-7B & Mistral-7B & Qwen2-7B & Llama-3-8B & Flan-T5-XXL & Granite-13B & Llama-2-13B & Mistral-8x7B & Llama-2-70B \\
\midrule
Drug (\%) & 0.1 & 0.2 & 0.0 & \cellcolor{gray}\textit{5.5} & 0.4 & 0.6 & 1.2 & 1.6 & 1.0 \\
Material (\%) & 0.3 & 0.4 & 0.0 & \cellcolor{gray}\textit{4.8} & 0.8 & 1.6 & 1.2 & 1.7 & 0.8 \\
\end{tabular} \\
\midrule
\begin{tabular}{@{}lccccccccc@{}}
& & \multicolumn{4}{c}{\textbf{Supervised Fine-tuning}} & \multicolumn{3}{c}{\textbf{Llamole}} \\
\cmidrule(l){3-6} \cmidrule(l){7-9}
& & Mistral-7B & Qwen2-7B & Llama-3-8B & Llama-3.1-8B & Mistral-7B & Qwen2-7B & Llama-3.1-8B & \\
\midrule
Drug (\%) & & 1.5 & 0.2 & 0.6 & 0.8 & 29.9 & 33.7 & \cellcolor{red!15}\textbf{35.1} & \\
Material (\%) & & 0.8 & 0.1 & 0.7 & 0.8 & 14.3 & \cellcolor{red!15}\textbf{17.9} & 17.6 & \\
\end{tabular} \\
\bottomrule
\end{tabular}%
}
\end{table}

(3) \textbf{Larger models without domain-specific adaptation do not necessarily perform better in molecular designs.} We calculate the average Pearson correlation coefficient between model size and molecular design metrics, yielding a value of 0.366, indicating a weak correlation (below 0.5) between size and performance. We also compare LLM performance with GraphGA, which has been shown to be simple yet powerful~\citep{gao2022sample,liu2024inverse}. Our observations confirm that GraphGA serves as a strong molecular design baseline, challenging most LLM models with ICL and SFT in generating molecules with precise multi-condition control.


\subsection{RQ2: Discussion on Controllable Molecular Generation}

\begin{figure}[t]
    \centering
    \vspace{-0.15in}
    \includegraphics[width=0.9\textwidth]{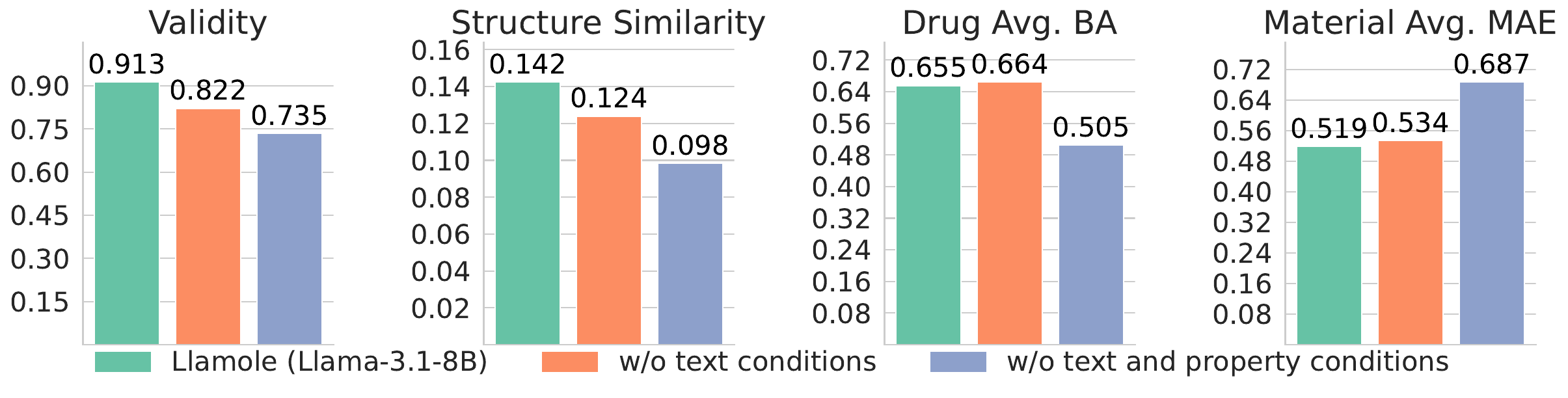}
    \caption{Ablation Studies for the Graph DiT Module in \method: First, we remove the text conditions from the input, i.e., $\mathbf{c}_\text{text}=\emptyset$. Next, we remove both text and property conditions, $\{c_i\}_i^M \cup \mathbf{c}_\text{text}$. There are learned embeddings that represent the ``null'' value for different conditions.
    }
    \label{fig:ablation-design}
\end{figure}
 
\subsubsection{Ablation Studies on LLM and Graph DiT Synergy}
We investigate the synergy effect of Graph DiT and LLM in \method for molecule controllability.
We first remove text conditions $\mathbf{c}_\text{text}$. In this case, Graph DiT uses a learned ``null'' embedding to represent the dropped condition $\mathbf{c}_\text{text}=\emptyset$.
Next, we remove the drug or material property conditions $\{c_i\}_i^M$ associated with the question.
Results in~\cref{fig:ablation-design} show that text instructions enhance the chemical structure understanding ability of Graph DiT, while \method leverages Graph DiT’s capabilities with property inputs to generate molecules with desirable properties.

\begin{figure}[!t]
    \centering
    \vspace{-0.1in}    
    \includegraphics[width=\textwidth]{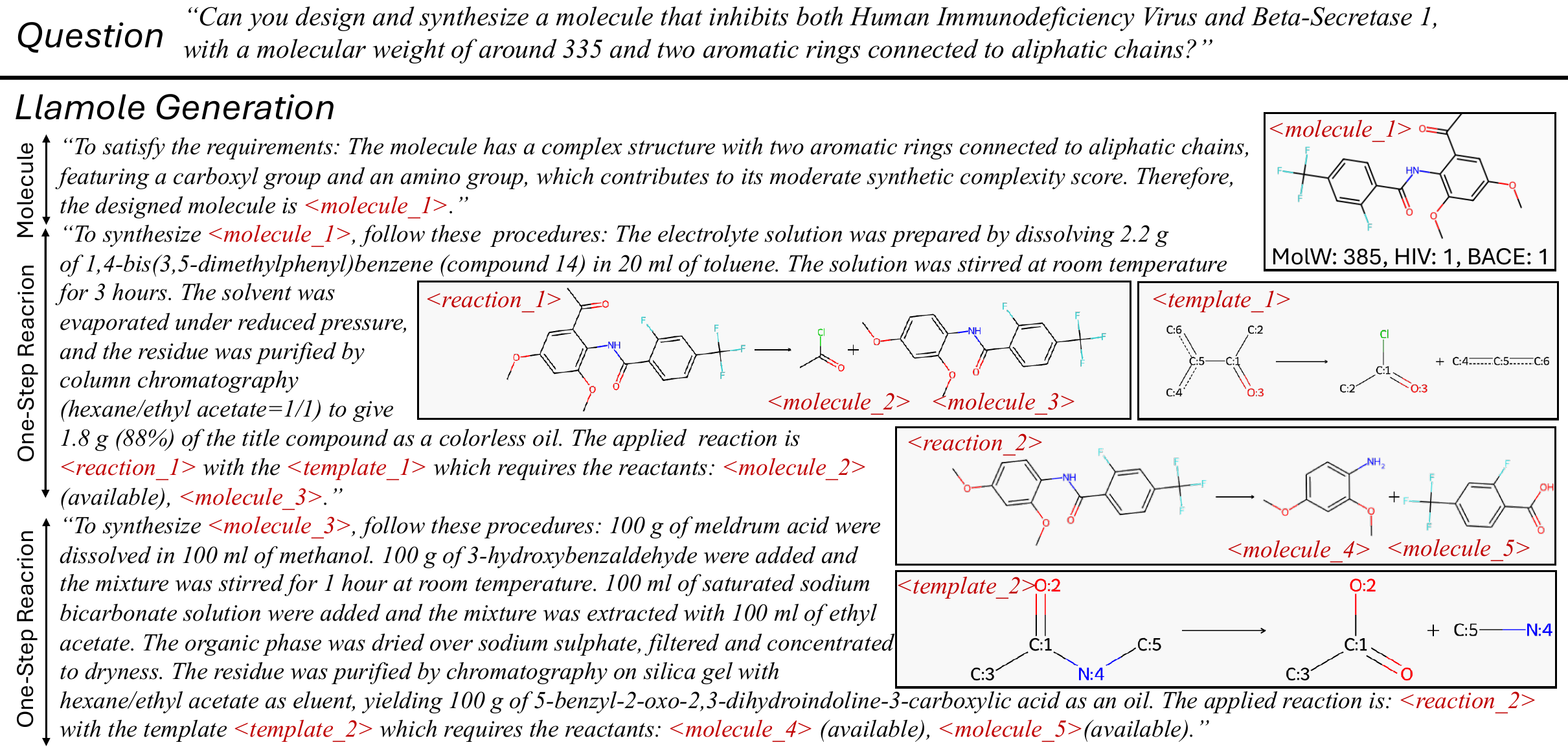}
    \vspace{-0.2in}
    \caption{Interleaved generation with the base Qwen2-7B: Red indicates positions where molecules and reactions (with templates) are generated, forming three parts. The properties of the designed molecules are obtained from the oracle. Reference and other LLM responses are shown in~\cref{fig:case1_more}.}
    \label{fig:case1}
    \vspace{-0.25in}
\end{figure}

\subsubsection{Case Studies for Property and Structure Controllability}

In~\cref{fig:case1}, \method can design a satisfactory molecule that meets both functional and structural constraints. Functionally, the oracle function confirms that the properties of BACE and HIV align with the criteria. Structurally, all key criteria are satisfied, including molecular weight, ``two aromatic rings,'' and ``connected to aliphatic chains.'' \method also adds details for structure design, such as a carboxyl (\ce{-COOH}) group and an amino group (\ce{-NH2}). While the amino group is present in the structure, it is connected to the carbonyl group (\ce{-C(=O)-}) instead of the carboxyl group. This subtle difference may require precise control based on the text condition. More results are in~\cref{sec:add-case-study}.

\subsection{RQ3: Discussion on Retrosynthetic Planning}

\begin{wrapfigure}{r}{0.5\textwidth}
    \centering
    \includegraphics[width=0.5\textwidth]{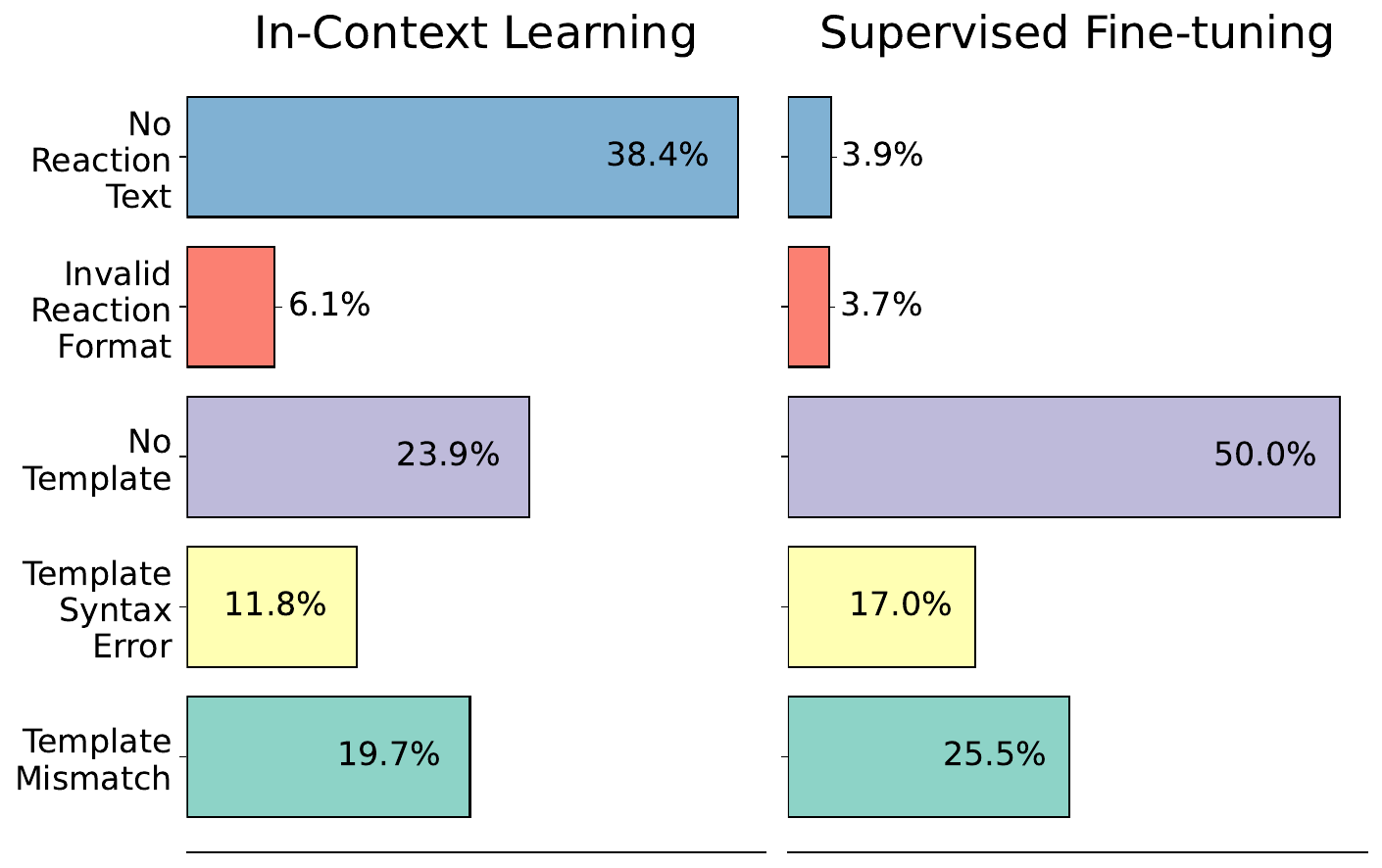}
    \caption{Error Analysis in Reaction Generation}
    \label{fig:llm-reaction-error}
    \vspace{-0.35in}
\end{wrapfigure}
Retrosynthesis challenges LLMs in two aspects: (1) one-step reaction generation and (2) multi-step planning. \cref{tab:retro-success-ratio} highlights the weaknesses of LLMs with ICL and SFT in overall planning ability and the promise of \method. 
We examine the failure reasons in LLMs and the synergy between the GNN and LLMs to avoid them.

\subsubsection{One-step Reaction Generation}
We conduct error analysis for LLMs in reaction generation. Results in~\cref{fig:llm-reaction-error} average performance across all LLMs using ICL or SFT methods. We identify five types of errors related to instruction adherence, format compliance, and template matching. We find that LLMs using ICL frequently fail to follow instructions for generating reactions in text format, with a high probability (68.4\%) of not producing valid formats and templates. In contrast, LLMs with SFT reduce this probability to 57.6\%. However, neither ICL nor SFT guarantees that the templates are correct or match the generated reactions.
In comparison, \method avoids these errors by using GNN predictors, which estimate probabilities for over 300K templates derived from USPTO reactions. This enables \method to apply templates directly to derive reactions in retrosynthesis, avoiding hallucination.


\subsubsection{Multi-step Retrosynthetic Planning}
From the success cases in~\cref{tab:retro-success-ratio}, we find that 96.40\% of 777 success cases in ICL-adapted LLMs and 94.14\% of 324 success cases in SFT-adapted LLMs arise from one-step reaction generation. However, not all designed molecules can be synthesized via one-step reactions. Compared to LLMs, \method achieves over 10K success cases, with 40.48\% resulting from two or more steps.
\cref{fig:case1} illustrates a two-step planning case for the designed molecule. The generation interleaves reaction conditions and specific formulas based on the template in both steps. 

\begin{wraptable}{r}{0.45\textwidth}
    \centering
    \caption{Analysis of $J_\text{heuristics}$ and Planning Time on Material Questions}
    \label{tab:retro-search-anlaysis}
    \begin{adjustbox}{width=0.45\textwidth} 
        \begin{tabular}{lccc}
            \toprule
            \shortstack{Base \\ LLM} & Default & \shortstack{w/ Domain \\ Heuristics} & \shortstack{w/ Unlimited \\ Time} \\
            \midrule
            Llama-3.1 & 0.176 & 0.176 & 0.312 \\
            Mistral   & 0.143 & 0.147 & 0.273 \\
            Qwen2     & 0.179 & 0.181 & 0.273 \\
            \bottomrule
        \end{tabular}
    \end{adjustbox}
    \vspace{-0.15in}
\end{wraptable}
\method is influenced by two factors for retrosynthesis: (1) the size of the search space and (2) the quality of the cost $J_\text{heuristic}$. The results reported in~\cref{tab:retro-success-ratio} limited the total planning time to 30 seconds (on an A6000 card). We remove this time constraint and report comparisons for material tasks in~\cref{tab:retro-search-anlaysis}. We find that success rates for all base LLMs significantly improve, but this comes at the cost of long inference time.
While there is a trade-off between efficiency and effectiveness, extending response time by a few minutes is often acceptable to improve the success rate of finding synthesis paths for a single designed molecule. In~\cref{tab:retro-search-anlaysis}, we also compare the $J_\text{heuristics}$ designed by LLMs (default) with the domain model trained from~\cite{chen2020retro}. we find that LLMs are competitive with these domain models in providing the cost function for A*, contrasting with previous observations where LLMs struggled with retrosynthetic planning.


\section{Related Work}
Since the emergence of ChatGPT~\citep{achiam2023gpt}, LLMs~\citep{dubey2024llama} have become foundation models for text-based problems and are revolutionizing domains like vision and speech~\citep{dong2023dreamllm,wu2024nextgpt}. These advancements extend to chemistry, biology, and material sciences, focusing on molecules~\citep{guo2023can,jablonka202314,jin2023large}. Prior work explores LLMs in molecular generation, property prediction, and one-step reaction prediction in retrosynthesis~\citep{guo2023can,jablonka202314}. A key lesson is the limitation of LLMs in sequential modeling of molecules (e.g., SMILES or SELFIES)~\citep{guo2023can}. Multimodal LMs have been developed in combination with GNNs~\citep{zhao2023gimlet,liu2023molca}, but they have not been scaled up to leverage LLMs with at least 7 billion parameters.
Additionally, LLMs struggle with planning tasks~\citep{kambhampati2024llms}, which are essential for retrosynthesis. We address these issues using graph-text multimodal LLMs, augmented by A* for efficient planning.

Domain-specific molecular design methods have evolved from sequential models~\citep{segler2018generating} to graph diffusion models~\citep{jo2022score,vignac2022digress,liu2024inverse}. Studies show that older graph-based methods like GraphGA remain competitive~\citep{gao2022sample}. To incorporate property constraints, one can use molecular optimization approaches such as Bayesian optimization or REINFORCE~\citep{gao2022sample}, or employ diffusion models with or without predictor guidance~\citep{vignac2022digress,liu2024inverse}.
For synthesizable molecular design, prior work has focused on bottom-up methods~\citep{gao2021amortized,sun2024syntax}. These methods explore a chemical space defined by a discrete action space of reaction templates and purchasable starting materials, which may limit flexibility. Thus, retrosynthesis algorithms~\citep{chen2020retro} are also studied as separate solutions to find synthesis routes for generated molecules in a top-down manner.



\section{Conclusion}
We have presented the first graph-text MLLM, \method, for multi-conditional molecular generation and retrosynthetic planning. By integrating a base LLM with specialized graph modules, \method interleaved the generation of text, molecular graphs, and reactions, enabling controllable and synthesizable designs. Extensive benchmarking against 14 LLMs revealed their limitations in controlling molecular structures and planning synthesis routes. In contrast, \method significantly outperformed these LLMs. These findings underscored the value of multimodal approaches in molecular discovery and highlighted \method's potential to connect text and chemical structures. The new benchmarking dataset also laid the groundwork for future MLLM research in molecular applications.



\bibliography{reference}

\begin{thebibliography}{47}
\providecommand{\natexlab}[1]{#1}
\providecommand{\url}[1]{\texttt{#1}}
\expandafter\ifx\csname urlstyle\endcsname\relax
  \providecommand{\doi}[1]{doi: #1}\else
  \providecommand{\doi}{doi: \begingroup \urlstyle{rm}\Url}\fi

\bibitem[Abdelaziz et~al.(2024)Abdelaziz, Basu, Agarwal, Kumaravel, Stallone, Panda, Rizk, Bhargav, Crouse, Gunasekara, et~al.]{abdelaziz2024granite}
Ibrahim Abdelaziz, Kinjal Basu, Mayank Agarwal, Sadhana Kumaravel, Matthew Stallone, Rameswar Panda, Yara Rizk, GP~Bhargav, Maxwell Crouse, Chulaka Gunasekara, et~al.
\newblock Granite-function calling model: Introducing function calling abilities via multi-task learning of granular tasks.
\newblock \emph{arXiv preprint arXiv:2407.00121}, 2024.

\bibitem[Achiam et~al.(2023)Achiam, Adler, Agarwal, Ahmad, Akkaya, Aleman, Almeida, Altenschmidt, Altman, Anadkat, et~al.]{achiam2023gpt}
Josh Achiam, Steven Adler, Sandhini Agarwal, Lama Ahmad, Ilge Akkaya, Florencia~Leoni Aleman, Diogo Almeida, Janko Altenschmidt, Sam Altman, Shyamal Anadkat, et~al.
\newblock Gpt-4 technical report.
\newblock \emph{arXiv preprint arXiv:2303.08774}, 2023.

\bibitem[Austin et~al.(2021)Austin, Johnson, Ho, Tarlow, and Van Den~Berg]{austin2021structured}
Jacob Austin, Daniel~D Johnson, Jonathan Ho, Daniel Tarlow, and Rianne Van Den~Berg.
\newblock Structured denoising diffusion models in discrete state-spaces.
\newblock \emph{Advances in Neural Information Processing Systems}, 34:\penalty0 17981--17993, 2021.

\bibitem[Beltagy et~al.(2019)Beltagy, Lo, and Cohan]{beltagy2019scibert}
Iz~Beltagy, Kyle Lo, and Arman Cohan.
\newblock Scibert: A pretrained language model for scientific text.
\newblock \emph{arXiv preprint arXiv:1903.10676}, 2019.

\bibitem[Chen et~al.(2020)Chen, Li, Dai, and Song]{chen2020retro}
Binghong Chen, Chengtao Li, Hanjun Dai, and Le~Song.
\newblock Retro*: learning retrosynthetic planning with neural guided a* search.
\newblock In \emph{International conference on machine learning}, pages 1608--1616. PMLR, 2020.

\bibitem[Chung et~al.(2024)Chung, Hou, Longpre, Zoph, Tay, Fedus, Li, Wang, Dehghani, Brahma, et~al.]{chung2024scaling}
Hyung~Won Chung, Le~Hou, Shayne Longpre, Barret Zoph, Yi~Tay, William Fedus, Yunxuan Li, Xuezhi Wang, Mostafa Dehghani, Siddhartha Brahma, et~al.
\newblock Scaling instruction-finetuned language models.
\newblock \emph{Journal of Machine Learning Research}, 25\penalty0 (70):\penalty0 1--53, 2024.

\bibitem[Coley et~al.(2018)Coley, Rogers, Green, and Jensen]{coley2018scscore}
Connor~W Coley, Luke Rogers, William~H Green, and Klavs~F Jensen.
\newblock Scscore: synthetic complexity learned from a reaction corpus.
\newblock \emph{Journal of chemical information and modeling}, 58\penalty0 (2):\penalty0 252--261, 2018.

\bibitem[Coley et~al.(2019)Coley, Green, and Jensen]{coley2019rdchiral}
Connor~W Coley, William~H Green, and Klavs~F Jensen.
\newblock Rdchiral: An rdkit wrapper for handling stereochemistry in retrosynthetic template extraction and application.
\newblock \emph{Journal of chemical information and modeling}, 59\penalty0 (6):\penalty0 2529--2537, 2019.

\bibitem[Dong et~al.(2023)Dong, Han, Peng, Qi, Ge, Yang, Zhao, Sun, Zhou, Wei, et~al.]{dong2023dreamllm}
Runpei Dong, Chunrui Han, Yuang Peng, Zekun Qi, Zheng Ge, Jinrong Yang, Liang Zhao, Jianjian Sun, Hongyu Zhou, Haoran Wei, et~al.
\newblock Dreamllm: Synergistic multimodal comprehension and creation.
\newblock \emph{arXiv preprint arXiv:2309.11499}, 2023.

\bibitem[Dubey et~al.(2024)Dubey, Jauhri, Pandey, Kadian, Al-Dahle, Letman, Mathur, Schelten, Yang, Fan, et~al.]{dubey2024llama}
Abhimanyu Dubey, Abhinav Jauhri, Abhinav Pandey, Abhishek Kadian, Ahmad Al-Dahle, Aiesha Letman, Akhil Mathur, Alan Schelten, Amy Yang, Angela Fan, et~al.
\newblock The llama 3 herd of models.
\newblock \emph{arXiv preprint arXiv:2407.21783}, 2024.

\bibitem[Ertl and Schuffenhauer(2009)]{ertl2009estimation}
Peter Ertl and Ansgar Schuffenhauer.
\newblock Estimation of synthetic accessibility score of drug-like molecules based on molecular complexity and fragment contributions.
\newblock \emph{Journal of cheminformatics}, 1:\penalty0 1--11, 2009.

\bibitem[Gao et~al.(2021)Gao, Mercado, and Coley]{gao2021amortized}
Wenhao Gao, Roc{\'\i}o Mercado, and Connor~W Coley.
\newblock Amortized tree generation for bottom-up synthesis planning and synthesizable molecular design.
\newblock \emph{arXiv preprint arXiv:2110.06389}, 2021.

\bibitem[Gao et~al.(2022)Gao, Fu, Sun, and Coley]{gao2022sample}
Wenhao Gao, Tianfan Fu, Jimeng Sun, and Connor Coley.
\newblock Sample efficiency matters: a benchmark for practical molecular optimization.
\newblock \emph{Advances in neural information processing systems}, 35:\penalty0 21342--21357, 2022.

\bibitem[Guo et~al.(2023)Guo, Nan, Liang, Guo, Chawla, Wiest, Zhang, et~al.]{guo2023can}
Taicheng Guo, Bozhao Nan, Zhenwen Liang, Zhichun Guo, Nitesh Chawla, Olaf Wiest, Xiangliang Zhang, et~al.
\newblock What can large language models do in chemistry? a comprehensive benchmark on eight tasks.
\newblock \emph{Advances in Neural Information Processing Systems}, 36:\penalty0 59662--59688, 2023.

\bibitem[Ho and Salimans(2022)]{ho2022classifier}
Jonathan Ho and Tim Salimans.
\newblock Classifier-free diffusion guidance.
\newblock \emph{arXiv preprint arXiv:2207.12598}, 2022.

\bibitem[Hu et~al.(2021)Hu, Shen, Wallis, Allen-Zhu, Li, Wang, Wang, and Chen]{hu2021lora}
Edward~J Hu, Yelong Shen, Phillip Wallis, Zeyuan Allen-Zhu, Yuanzhi Li, Shean Wang, Lu~Wang, and Weizhu Chen.
\newblock Lora: Low-rank adaptation of large language models.
\newblock \emph{arXiv preprint arXiv:2106.09685}, 2021.

\bibitem[Jablonka et~al.(2023)Jablonka, Ai, Al-Feghali, Badhwar, Bocarsly, Bran, Bringuier, Brinson, Choudhary, Circi, et~al.]{jablonka202314}
Kevin~Maik Jablonka, Qianxiang Ai, Alexander Al-Feghali, Shruti Badhwar, Joshua~D Bocarsly, Andres~M Bran, Stefan Bringuier, L~Catherine Brinson, Kamal Choudhary, Defne Circi, et~al.
\newblock 14 examples of how llms can transform materials science and chemistry: a reflection on a large language model hackathon.
\newblock \emph{Digital Discovery}, 2\penalty0 (5):\penalty0 1233--1250, 2023.

\bibitem[Jiang et~al.(2023)Jiang, Sablayrolles, Mensch, Bamford, Chaplot, Casas, Bressand, Lengyel, Lample, Saulnier, et~al.]{jiang2023mistral}
Albert~Q Jiang, Alexandre Sablayrolles, Arthur Mensch, Chris Bamford, Devendra~Singh Chaplot, Diego de~las Casas, Florian Bressand, Gianna Lengyel, Guillaume Lample, Lucile Saulnier, et~al.
\newblock Mistral 7b.
\newblock \emph{arXiv preprint arXiv:2310.06825}, 2023.

\bibitem[Jin et~al.(2023)Jin, Liu, Han, Jiang, Ji, and Han]{jin2023large}
Bowen Jin, Gang Liu, Chi Han, Meng Jiang, Heng Ji, and Jiawei Han.
\newblock Large language models on graphs: A comprehensive survey.
\newblock \emph{arXiv preprint arXiv:2312.02783}, 2023.

\bibitem[Jo et~al.(2022)Jo, Lee, and Hwang]{jo2022score}
Jaehyeong Jo, Seul Lee, and Sung~Ju Hwang.
\newblock Score-based generative modeling of graphs via the system of stochastic differential equations.
\newblock In \emph{International Conference on Machine Learning}, volume 162, pages 10362--10383. PMLR, 2022.

\bibitem[Kambhampati et~al.(2024)Kambhampati, Valmeekam, Guan, Stechly, Verma, Bhambri, Saldyt, and Murthy]{kambhampati2024llms}
Subbarao Kambhampati, Karthik Valmeekam, Lin Guan, Kaya Stechly, Mudit Verma, Siddhant Bhambri, Lucas Saldyt, and Anil Murthy.
\newblock Llms can't plan, but can help planning in llm-modulo frameworks.
\newblock \emph{arXiv preprint arXiv:2402.01817}, 2024.

\bibitem[Kim et~al.(2021)Kim, Chen, Cheng, Gindulyte, He, He, Li, Shoemaker, Thiessen, Yu, et~al.]{kim2021pubchem}
Sunghwan Kim, Jie Chen, Tiejun Cheng, Asta Gindulyte, Jia He, Siqian He, Qingliang Li, Benjamin~A Shoemaker, Paul~A Thiessen, Bo~Yu, et~al.
\newblock Pubchem in 2021: new data content and improved web interfaces.
\newblock \emph{Nucleic acids research}, 49\penalty0 (D1):\penalty0 D1388--D1395, 2021.

\bibitem[Liu et~al.(2022)Liu, Zhao, Xu, Luo, and Jiang]{liu2022graph}
Gang Liu, Tong Zhao, Jiaxin Xu, Tengfei Luo, and Meng Jiang.
\newblock Graph rationalization with environment-based augmentations.
\newblock In \emph{Proceedings of the 28th ACM SIGKDD Conference on Knowledge Discovery and Data Mining}, pages 1069--1078, 2022.

\bibitem[Liu et~al.(2023{\natexlab{a}})Liu, Zhao, Inae, Luo, and Jiang]{liu2023semi}
Gang Liu, Tong Zhao, Eric Inae, Tengfei Luo, and Meng Jiang.
\newblock Semi-supervised graph imbalanced regression.
\newblock In \emph{Proceedings of the 29th ACM SIGKDD Conference on Knowledge Discovery and Data Mining}, pages 1453--1465, 2023{\natexlab{a}}.

\bibitem[Liu et~al.(2024{\natexlab{a}})Liu, Inae, Luo, and Jiang]{liu2024rationalizing}
Gang Liu, Eric Inae, Tengfei Luo, and Meng Jiang.
\newblock Rationalizing graph neural networks with data augmentation.
\newblock \emph{ACM Transactions on Knowledge Discovery from Data}, 18\penalty0 (4):\penalty0 1--23, 2024{\natexlab{a}}.

\bibitem[Liu et~al.(2024{\natexlab{b}})Liu, Inae, Zhao, Xu, Luo, and Jiang]{liu2024data}
Gang Liu, Eric Inae, Tong Zhao, Jiaxin Xu, Tengfei Luo, and Meng Jiang.
\newblock Data-centric learning from unlabeled graphs with diffusion model.
\newblock \emph{Advances in neural information processing systems}, 36, 2024{\natexlab{b}}.

\bibitem[Liu et~al.(2024{\natexlab{c}})Liu, Xu, Luo, and Jiang]{liu2024inverse}
Gang Liu, Jiaxin Xu, Tengfei Luo, and Meng Jiang.
\newblock Inverse molecular design with multi-conditional diffusion guidance.
\newblock \emph{arXiv preprint arXiv:2401.13858}, 2024{\natexlab{c}}.

\bibitem[Liu et~al.(2023{\natexlab{b}})Liu, Li, Luo, Fei, Cao, Kawaguchi, Wang, and Chua]{liu2023molca}
Zhiyuan Liu, Sihang Li, Yanchen Luo, Hao Fei, Yixin Cao, Kenji Kawaguchi, Xiang Wang, and Tat-Seng Chua.
\newblock Molca: Molecular graph-language modeling with cross-modal projector and uni-modal adapter.
\newblock In \emph{Proceedings of the 2023 Conference on Empirical Methods in Natural Language Processing}, pages 15623--15638, 2023{\natexlab{b}}.

\bibitem[Lowe(2017)]{Lowe2017}
Daniel Lowe.
\newblock {Chemical reactions from US patents (1976 Sep2016)}.
\newblock 6 2017.
\newblock \doi{10.6084/m9.figshare.5104873.v1}.
\newblock URL \url{https://figshare.com/articles/dataset/Chemical_reactions_from_US_patents_1976-Sep2016_/5104873}.

\bibitem[Ma and Luo(2020)]{ma2020pi1m}
Ruimin Ma and Tengfei Luo.
\newblock Pi1m: a benchmark database for polymer informatics.
\newblock \emph{Journal of Chemical Information and Modeling}, 60\penalty0 (10):\penalty0 4684--4690, 2020.

\bibitem[Otsuka et~al.(2011)Otsuka, Kuwajima, Hosoya, Xu, and Yamazaki]{otsuka2011polyinfo}
Shingo Otsuka, Isao Kuwajima, Junko Hosoya, Yibin Xu, and Masayoshi Yamazaki.
\newblock Polyinfo: Polymer database for polymeric materials design.
\newblock In \emph{2011 International Conference on Emerging Intelligent Data and Web Technologies}, pages 22--29. IEEE, 2011.

\bibitem[Ouyang et~al.(2022)Ouyang, Wu, Jiang, Almeida, Wainwright, Mishkin, Zhang, Agarwal, Slama, Ray, et~al.]{ouyang2022training}
Long Ouyang, Jeffrey Wu, Xu~Jiang, Diogo Almeida, Carroll Wainwright, Pamela Mishkin, Chong Zhang, Sandhini Agarwal, Katarina Slama, Alex Ray, et~al.
\newblock Training language models to follow instructions with human feedback.
\newblock \emph{Advances in neural information processing systems}, 35:\penalty0 27730--27744, 2022.

\bibitem[Radford et~al.(2021)Radford, Kim, Hallacy, Ramesh, Goh, Agarwal, Sastry, Askell, Mishkin, Clark, et~al.]{radford2021learning}
Alec Radford, Jong~Wook Kim, Chris Hallacy, Aditya Ramesh, Gabriel Goh, Sandhini Agarwal, Girish Sastry, Amanda Askell, Pamela Mishkin, Jack Clark, et~al.
\newblock Learning transferable visual models from natural language supervision.
\newblock In \emph{International conference on machine learning}, pages 8748--8763. PMLR, 2021.

\bibitem[Rogers and Hahn(2010)]{rogers2010extended}
David Rogers and Mathew Hahn.
\newblock Extended-connectivity fingerprints.
\newblock \emph{Journal of chemical information and modeling}, 50\penalty0 (5):\penalty0 742--754, 2010.

\bibitem[Segler et~al.(2018)Segler, Kogej, Tyrchan, and Waller]{segler2018generating}
Marwin~HS Segler, Thierry Kogej, Christian Tyrchan, and Mark~P Waller.
\newblock Generating focused molecule libraries for drug discovery with recurrent neural networks.
\newblock \emph{ACS central science}, 4\penalty0 (1):\penalty0 120--131, 2018.

\bibitem[Sterling and Irwin(2015)]{sterling2015zinc}
Teague Sterling and John~J Irwin.
\newblock Zinc 15--ligand discovery for everyone.
\newblock \emph{Journal of chemical information and modeling}, 55\penalty0 (11):\penalty0 2324--2337, 2015.

\bibitem[Sun et~al.(2024)Sun, Lo, Gao, Guo, Thost, Chen, Coley, and Matusik]{sun2024syntax}
Michael Sun, Alston Lo, Wenhao Gao, Minghao Guo, Veronika Thost, Jie Chen, Connor Coley, and Wojciech Matusik.
\newblock Syntax-guided procedural synthesis of molecules.
\newblock \emph{arXiv preprint arXiv:2409.05873}, 2024.

\bibitem[Swanson et~al.(2024)Swanson, Walther, Leitz, Mukherjee, Wu, Shivnaraine, and Zou]{swanson2024admet}
Kyle Swanson, Parker Walther, Jeremy Leitz, Souhrid Mukherjee, Joseph~C Wu, Rabindra~V Shivnaraine, and James Zou.
\newblock Admet-ai: a machine learning admet platform for evaluation of large-scale chemical libraries.
\newblock \emph{Bioinformatics}, 40\penalty0 (7):\penalty0 btae416, 2024.

\bibitem[Thornton et~al.(2012)Thornton, Robeson, Freeman, and Uhlmann]{thornton2012polymer}
A~Thornton, L~Robeson, B~Freeman, and D~Uhlmann.
\newblock Polymer gas separation membrane database, 2012.
\newblock URL \url{https://research.csiro.au/virtualscreening/membrane-database-polymer-gas-separation-membranes/}.

\bibitem[Vignac et~al.(2022)Vignac, Krawczuk, Siraudin, Wang, Cevher, and Frossard]{vignac2022digress}
Clement Vignac, Igor Krawczuk, Antoine Siraudin, Bohan Wang, Volkan Cevher, and Pascal Frossard.
\newblock Digress: Discrete denoising diffusion for graph generation.
\newblock \emph{arXiv preprint arXiv:2209.14734}, 2022.

\bibitem[Weininger(1988)]{weininger1988smiles}
David Weininger.
\newblock Smiles, a chemical language and information system. 1. introduction to methodology and encoding rules.
\newblock \emph{Journal of chemical information and computer sciences}, 28\penalty0 (1):\penalty0 31--36, 1988.

\bibitem[Wu et~al.(2024)Wu, Fei, Qu, Ji, and Chua]{wu2024nextgpt}
Shengqiong Wu, Hao Fei, Leigang Qu, Wei Ji, and Tat-Seng Chua.
\newblock {NE}xt-{GPT}: Any-to-any multimodal {LLM}.
\newblock In \emph{Forty-first International Conference on Machine Learning}, 2024.
\newblock URL \url{https://openreview.net/forum?id=NZQkumsNlf}.

\bibitem[Wu et~al.(2018)Wu, Ramsundar, Feinberg, Gomes, Geniesse, Pappu, Leswing, and Pande]{wu2018moleculenet}
Zhenqin Wu, Bharath Ramsundar, Evan~N Feinberg, Joseph Gomes, Caleb Geniesse, Aneesh~S Pappu, Karl Leswing, and Vijay Pande.
\newblock Moleculenet: a benchmark for molecular machine learning.
\newblock \emph{Chemical science}, 9\penalty0 (2):\penalty0 513--530, 2018.

\bibitem[Xu et~al.(2018)Xu, Hu, Leskovec, and Jegelka]{xu2018powerful}
Keyulu Xu, Weihua Hu, Jure Leskovec, and Stefanie Jegelka.
\newblock How powerful are graph neural networks?
\newblock \emph{arXiv preprint arXiv:1810.00826}, 2018.

\bibitem[Yang et~al.(2024)Yang, Yang, Hui, Zheng, Yu, Zhou, Li, Li, Liu, Huang, et~al.]{yang2024qwen2}
An~Yang, Baosong Yang, Binyuan Hui, Bo~Zheng, Bowen Yu, Chang Zhou, Chengpeng Li, Chengyuan Li, Dayiheng Liu, Fei Huang, et~al.
\newblock Qwen2 technical report.
\newblock \emph{arXiv preprint arXiv:2407.10671}, 2024.

\bibitem[Zdrazil et~al.(2024)Zdrazil, Felix, Hunter, Manners, Blackshaw, Corbett, de~Veij, Ioannidis, Lopez, Mosquera, et~al.]{zdrazil2024chembl}
Barbara Zdrazil, Eloy Felix, Fiona Hunter, Emma~J Manners, James Blackshaw, Sybilla Corbett, Marleen de~Veij, Harris Ioannidis, David~Mendez Lopez, Juan~F Mosquera, et~al.
\newblock The chembl database in 2023: a drug discovery platform spanning multiple bioactivity data types and time periods.
\newblock \emph{Nucleic acids research}, 52\penalty0 (D1):\penalty0 D1180--D1192, 2024.

\bibitem[Zhao et~al.(2023)Zhao, Liu, Chang, Xu, Fu, Deng, Kong, and Liu]{zhao2023gimlet}
Haiteng Zhao, Shengchao Liu, Ma~Chang, Hannan Xu, Jie Fu, Zhihong Deng, Lingpeng Kong, and Qi~Liu.
\newblock Gimlet: A unified graph-text model for instruction-based molecule zero-shot learning.
\newblock \emph{Advances in Neural Information Processing Systems}, 36:\penalty0 5850--5887, 2023.

\end{thebibliography}
\bibliographystyle{plainnat}

\newpage
\tableofcontents
\newpage
\appendix
\section{Additional Details for \method}\label{sec:add-method}

\subsection{Details of Special Tokens}

In total, there are nine special tokens divided into three groups. These tokens augment the word vocabulary $\mathcal{W}$, enabling flexible control of the generation flow:

\begin{itemize}
    \item Trigger and Query tokens: \texttt{<design\_start>}, \texttt{<design\_body>}, \texttt{<design\_end>}, \texttt{<retro\_start>}, \texttt{<retro\_body>}, \texttt{<retro\_end>}
    \item Molecule token: \texttt{<molecule>}
    \item Callback tokens: \texttt{<callback\_start>}, \texttt{<callback\_end>}
\end{itemize}

The tokens \texttt{<design\_start>} and \texttt{<retro\_start>} switch between the LLM and the Graph DiT or GNN, respectively. The tokens \texttt{<design\_body>} and \texttt{<retro\_body>} serve as query tokens, repeated eight times. After tokenization, the LLM takes their embeddings as input and outputs a vector from the last layer. The tokens \texttt{<design\_end>} and \texttt{<retro\_end>} indicate the end of these switches.

The \texttt{<molecule>} token marks the position of the molecular graph where the graph encoder is applied. In the instruction dataset, the segment ``\texttt{<mol\_start>}SMILES\texttt{<mol\_end>}'' denotes the position and identity of the molecule. SMILES will be converted to molecular graphs using RDKit, and this segment will be replaced by the \texttt{<molecule>} token for \method inputs.

Finally, callback tokens control the LLM to generate backup results as complements to the specialized graph modules. For instance, if the Graph DiT fails to produce a valid molecule, the base LLM can generate an alternative, regardless of validity.

\subsection{Details of LLM-based A* Heuristics}

\method models $J_\text{heuristics}$ in A* search as a multi-choice problem, filling in information from the molecule node, its parent reaction nodes and siblings using the template below. Parameters such as step, reaction template, and reactants are optional.

\begin{lstlisting} 
Estimate remaining steps for the target {smiles} given the following parameters: 
Current step {step}, 
Current template: {template}, 
Reactants: {reactants}. 
Consider the following factors: 
1. Intermediate complexity 
2. Reagent availability 
3. Side reactions 
4. Stereochemistry challenges.
\end{lstlisting}

Using this question to estimate remaining steps, we input the text into the base LLM and formulate five choices with corresponding scores:

\begin{lstlisting}
A. All readily available // Score: 0
B. Some commercial, some need 1-2 steps // Score: 1
C. Mix of commercial and multi-step synthesis // Score: 2.5
D. Mostly require complex synthesis // Score: 4.5
E. All require extensive multi-step synthesis // Score: 7
\end{lstlisting}

The LLM outputs logits for the next token, which we average for each choice to obtain overall probabilities. The $J_\text{heuristics}$ is calculated as the weighted score using these probabilities.

\section{Additional Benchmarking and Datasets Details}\label{sec:add-benchmark}

We collect small drug molecules from PubChem~\citep{kim2021pubchem}, MoleculeNet~\citep{wu2018moleculenet}, ChEMBL~\citep{zdrazil2024chembl}, and ZINC~\citep{sterling2015zinc}. 
Polymers are macromolecules with one repeating unit called monomers. We collect polymers from PI1M~\citep{ma2020pi1m}, the Membrane Society of Australia (MSA)~\citep{thornton2012polymer}, and others~\citep{liu2024data}.
Additionally, we collect 3.8 million patent chemical reactions with descriptions from USPTO~\citep{Lowe2017}, spanning from 1976 to 2016.

\subsection{Details of Quality Control}

After collecting molecules and polymers from various sources, we deduplicate and merge the label information for identical molecules. We use RDKit to obtain canonical SMILES. For small molecules, we calculate the first 14 characters of the InChIKey as the unique identifier, while for polymers, where the polymerization point is represented by ``*'', we use the canonical SMILES directly.

For drug-like small molecules, we apply the following rules to filter out alert structures, known as the Rule of Five (Ro5):

\begin{itemize}
    \item Molecular Weight (MW): Must be $\leq$ 500 Da.
    \item Hydrogen Bond Acceptors (HBA): Must not exceed 10.
    \item Hydrogen Bond Donors (HBD): Must not exceed 5.
    \item LogP: Must be $\leq$ 5, indicating lipophilicity.
\end{itemize}

A molecule passes the Ro5 test if at least three of these four conditions are met, indicating potential oral bioavailability.

We also apply 15 filter rules from the RDKit package, including the following from the FilterCatalogs Class: BRENK, CHEMBL, CHEMBL\_BMS, CHEMBL\_Dundee, CHEMBL\_Glaxo, CHEMBL\_Inpharmatica, CHEMBL\_LINT, CHEMBL\_MLSMR, CHEMBL\_SureChEMBL, NIH, PAINS, PAINS\_A, PAINS\_B, PAINS\_C, and ZINC.

\subsection{Details on the Creation of MolQA}\label{sec:add-create-molqa}

\subsubsection{Creation of Synthesis Routes}\label{sec:add-create-synthesis-route}

The USPTO has 3.7 million reactions. There are approximately 1.3 million unique product molecules. The purchasable compounds come from the Enamine Building Block (June 2024 version), supplemented with other common ions and starting materials, totaling around 1.3 million. We check each product from USPTO as a target molecule in the retrosynthesis task, exploring whether they can be synthesized using existing USPTO reactions through depth-first search (DFS). Ultimately, we identify about 139K target molecules with synthesis routes, supporting the creation of MolQA.

Since there are no polymerization reactions, we consider only monomer structures by replacing the * point with hydrogen. Among the 139K small molecules with synthesis routes, 2196 fit the monomer structures and serve as target molecules for polymer retrosynthesis. The length of synthesis routes ranges from 1 to 10. For each length of the routes, we split half of the molecules into the testing set, with a maximum of 3000, while the remainder is retained in the training set.

It results in around 11K routes (750 for materials and 9986 for drugs) for testing and 126K target molecules for training. 

\subsubsection{Creation of Property Annotations}

We focus on eight benchmarking properties: three drug-related categorical properties~\citep{wu2018moleculenet}—(1) HIV virus replication inhibition (HIV), (2) blood-brain barrier permeability (BBBP), and (3) human $\beta$-secretase 1 inhibition (BACE)—and five continuous material properties~\citep{thornton2012polymer}—(4) CO$_2$ permeability (CO$_2$Perm), (5) N$_2$ permeability (N$_2$Perm), (6) O$_2$ permeability (O$_2$Perm), (7) fractional free volume (FFV), and (8) thermal conductivity (TC). 

First, we check existing sources for annotations of these properties. 
To enrich the label space, we use well-trained GNN models~\citep{liu2022graph} to generate confident pseudo-labels, following the method in~\citep{liu2023semi}. We collect all labeled data to train two supervised multi-task GIN models for drug and material property annotation. The GIN models employ rationalization techniques~\citep{liu2024rationalizing} to split the molecular graph into rationale and environment subgraphs in the latent space, predicting labels from the rationale subgraph. The confidence score is computed by combining the rationale subgraph with various environment subgraphs, using the reciprocal of prediction variance. We annotate properties when prediction confidence exceeds the median threshold.

\subsubsection{Creation of Text Data for Molecular Description}\label{sec:add-create-text-for-molecule}

In addition to property annotations, we consider structural and synthesis information of the molecules using RDKit and heuristic complexity estimation scores. First, for any molecule, we extract the following structural information:
\begin{itemize}
    \item \textbf{Scaffold:} Extracted scaffold from the molecule structure.
    \item \textbf{Molecular Weight:} Calculated using the molecular weight descriptor.
    \item \textbf{Number of Rings:} Total number of rings in the molecule.
    \item \textbf{Number of Aromatic Rings:} Total number of aromatic rings in the molecule.
    \item \textbf{Number of Aliphatic Rings:} Total number of aliphatic rings in the molecule.
    \item \textbf{Number of Rotatable Bonds:} Total number of rotatable bonds in the molecule.
    \item \textbf{Number of Hydrogen Bond Donors:} Total number of hydrogen bond donors.
    \item \textbf{Number of Hydrogen Bond Acceptors:} Total number of hydrogen bond acceptors.
\end{itemize}

Next, we compute the synthetic accessibility score (SAScore)~\citep{ertl2009estimation} and SCScore~\citep{coley2018scscore}. Based on this information, we use the following template:

\begin{lstlisting}
Generate a summary description that starts directly with "The molecule/polymer ..." based on the predicted chemical properties, synthetic complexity scores, and structural information for the molecule with SMILES: {{smiles}}. Use your own knowledge, focus on functions, and avoid using numbers, redundant words, or mentioning SMILES. Ensure the output sentence is complete and ends with a period. This is for Drug/Material Utility of a Molecule/Polymer:

The structural context of a molecule includes its scaffold, which is the core structure around which the molecule is built. Key structural features include the presence of aromatic rings, aliphatic chains, and common functional groups such as hydroxyl, carboxyl, and amino groups. The complexity of the molecule's structure can significantly influence its physical and chemical properties.
Scaffold: {{scaffold}}
Molecular Weight: {{mw}}
Number of Rings: {{num_rings}}
Number of Aromatic Rings: {{num_arom_rings}}
Number of Aliphatic Rings: {{num_aliph_rings}}
Number of Rotatable Bonds: {{num_rot_bonds}}
Number of Hydrogen Bond Donors: {{num_h_donors}}
Number of Hydrogen Bond Acceptors: {{num_h_acceptors}}

{utility_context}
{{properties}}
\end{lstlisting}

The pre-defined utility context for the small molecule is as follows:
\begin{lstlisting}
The drug utility of a molecule is assessed based on its potential to serve as a therapeutic agent. Key properties considered include pharmacokinetics, which encompasses absorption, distribution, metabolism, excretion (ADME), and toxicity. Bioactivity is another critical factor, measured by the molecule's ability to interact with biological targets, typically through binding affinity. Additionally, drug-likeness, which refers to the molecule's adherence to established rules such as Lipinski's Rule of Five, is essential. This rule evaluates molecular weight, hydrogen bond donors and acceptors, and lipophilicity to predict a molecule's suitability as an oral drug.
\end{lstlisting}

The pre-defined utility context for the polymer is as follows:
\begin{lstlisting}
The material utility of a molecule, particularly for creating polymeric materials, is evaluated based on properties like mechanical strength, flexibility, and thermal and electrical behavior. For polymer membranes used in gas separation, crucial factors include gas permeability, which determines the efficiency of gas diffusion, and chemical stability, ensuring resistance to degradation. Additionally, thermal properties such as melting point and thermal conductivity are vital, as they affect the material's performance under various temperature conditions. Electrical properties, such as conductivity and dielectric constant, may also be significant depending on the intended application.
\end{lstlisting}

For the property variable, we include the property name with values, as well as the minimum, maximum, and percentile among the labels in the template. We repeat all annotated properties in the property variable. The estimated synthesis complexity scores are included among them.

We also prompt Llama-3-70B to generate short responses of 50-70 words, producing a molecular description for each molecule based on its properties, structures, and synthesis estimation. If a molecule has a description from PubChem~\citep{kim2021pubchem}, we concatenate these descriptions.

The generated texts may not always be meaningful or valid. We can establish filter rules based on patterns observed in poorly generated texts to remove them. We then regenerate texts for these items. After several iterations, we obtain the final text data for molecular utility descriptions, improving overall text quality. We also apply this strategy to other steps that involves prompting LLMs for synthetic data creation.

\subsubsection{Creation of Question Answering Data}

After annotating molecular description texts from~\cref{sec:add-create-text-for-molecule}, we combine them with reaction descriptions, including the reaction formula and template from synthesis routes in~\cref{sec:add-create-synthesis-route}. This forms the answer data in a QA data pair.

Next, we prompt Llame-3-70B to generate questions for each answer based on the following template.

\begin{lstlisting}
I'm creating a question-answer dataset for LLM fine-tuning. 
The question is about designing a molecule/polymer with these properties: {property_info} and the following structure information: {structure_info}. 
The expected answer for the question is: {answer}
Generate a SINGLE question about designing and synthesizing such a molecule/polymer that meets these criteria: 
(1) Start with 'Question:'; (2) End with a question mark; 
(3) Sound natural; (4) Be diverse; (5) Avoid redundancy and introductory words (like 'Here is a question that meets the criteria:')
(6) Do not include the answer; (7) Do not include incorrect information.

Example questions: 
(1) How can I design and synthesize a molecule with X, Y, and Z properties? 
(2) What is the best way to create a polymer with X, Y, and Z characteristics? 
(3) How to design a molecule with X, Y, and Z features and synthesize it? 
(4) I want a molecule with X, Y properties and Z structures. Please design it and describe the synthesis path.
\end{lstlisting}

The template is applied to any answer with the corresponding structure, property information, and complete answer texts. 

\subsection{Details on the Creation of MolPair}\label{sec:add-create-molpair}

MolPair consists of two parts: reaction-text pairs and graph-text pairs. 
We curate reaction-text pairs from USPTO~\citep{Lowe2017}, pairing each reaction with its corresponding description of the reaction conditions. We first deduplicate product molecules in reactions, obtaining input data as the product molecule alongside the reaction condition texts. Next, we extract reaction templates from the reaction formula using rdchiral~\citep{coley2019rdchiral}, resulting in approximately 300K templates, which will serve as labels for predictions. Finally, we have approximately 1.6 million training examples.


For the graph-text pairs, we use small molecules and polymers from the multisource collection, excluding those in MolQA. We follow the same pipeline used to create property and text annotations for the MolQA data, focusing on broader properties that describe drug-related utility with 41 small molecule properties~\citep{swanson2024admet}. Besides the three used in MolQA, others include:
\begin{itemize}
    \item Toxicity and Safety: AMES, Carcinogens Lagunin, ClinTox, DILI, Skin Reaction, hERG
    \item Enzyme Interaction: CYP1A2 Veith, CYP2C19 Veith, CYP2C9 Substrate CarbonMangels, CYP2C9 Veith, CYP2D6 Substrate CarbonMangels, CYP2D6 Veith, CYP3A4 Substrate CarbonMangels, CYP3A4 Veith
    \item Absorption, Distribution, Metabolism, and Excretion (ADME): BBB Martins, Bioavailability Ma, Caco2 Wang, Clearance Hepatocyte AZ, Clearance Microsome AZ, HIA Hou, Half Life Obach, Hydration Free Energy FreeSolv, Lipophilicity AstraZeneca, PAMPA NCATS, PPBR AZ, Pgp Broccatelli, Solubility AqSolDB, VDss Lombardo
    \item Stress Response: SR-ARE, SR-ATAD5, SR-HSE, SR-MMP, SR-p53
    \item Nuclear Receptor Interaction: NR-AR-LBD, NR-AR, NR-AhR, NR-Aromatase, NR-ER-LBD, NR-ER, NR-PPAR-gamma
\end{itemize}

We describe polymeric material utility based on 14 polymer properties collected from~\citet{otsuka2011polyinfo}:
\begin{itemize}
    \item Thermal Properties: Melting temperature [°C]; Specific heat capacity at constant pressure (\( C_p \)) [cal/(g·°C)]; Specific heat capacity at constant volume (\( C_v \)) [cal/(g·°C)]; Thermal conductivity [W/(m·K)]
    \item Physical \& Thermodynamic Properties: Density [g/cm\(^3\)]; Fractional Free Volume (dimensionless); Radius of Gyration (\( R_g \)) [nm]
    \item Permeability Properties: Gas diffusion coefficient (\( D \)) [cm\(^2\)/s]; Gas permeability coefficient (\( P \)) [cm\(^3\) (STP)·cm/(cm\(^2\)·s·Pa)]; Oxygen (\( \text{O}_2 \)) Gas Permeability (Barrer); Nitrogen (\( \text{N}_2 \)) Gas Permeability (Barrer); Carbon Dioxide (\( \text{CO}_2 \)) Gas Permeability (Barrer)
    \item Solubility Properties: Gas solubility coefficient (\( S \)) [cm\(^3\) (STP)·cm/(cm\(^2\)·s·Pa)]
    \item Dielectric \& Optical Properties: Dielectric constant.
\end{itemize}
We train two multi-task GIN models based on the rationalization method~\citep{liu2022graph} using all existing labeled data for drug and material property prediction, respectively. We use these models to predict properties for millions of small molecules and polymers, retaining the top ten thousand predictions by confidence score for each property. These are then used to prompt Llama-3-70B to create molecular descriptions, using the same prompt template as in~\cref{sec:add-create-text-for-molecule}. Additionally, we apply the same strategy as in~\cref{sec:add-create-text-for-molecule} to annotate labels for the eight studied properties, which can serve as input for pretraining the multi-conditional Graph DiT. Finally, we have approximately 300K graph-text pairs for small molecules and 300K graph-text pairs for polymers.

\section{Additional Pre-training and Fine-tuning Details}\label{sec:add-pretrain}

We pre-train three graph models including Graph DiT~\citep{liu2024inverse} for multi-conditional molecular generation, a GIN-based GNN predictor for reaction template prediction, and a GIN-based graph encoder for molecule understanding~\citep{xu2018powerful}.

\subsection{Pre-training of Graph Diffusion Transformers}

Suppose the node has \( F_V \) categories and the edge has \( F_E \) categories (including non-bond). Graph DiT models the node token by concatenating all its edge configurations to other nodes. For each node \( \mathbf{x} \in \mathbb{R}^F \), we have \( F = F_V + N_G \times F_E \), where \( N_G \) denotes the graph size. This facilitates defining the transition matrix \( \mathbf{Q} \) for the joint distribution of nodes and edges~\citep{liu2024inverse}. Graph DiT uses Transformer layers, replacing layer normalization with adaptive layer normalization (AdaLN):
\[
\operatorname{AdaLN}\left(\mathbf{h}, \mathbf{c}\right) = \gamma_\theta(\mathbf{c}) \odot \frac{\mathbf{h}-\mu\left(\mathbf{h}\right)}{\sigma\left(\mathbf{h}\right)}+\beta_\theta(\mathbf{c}),
\]
where \( \mathbf{h} \) denotes the hidden state of \( \mathbf{x} \) and \( \mathbf{c} \) is the vector representing the input conditions.

Given multiple conditions with categorical, continuous properties, and text, Graph DiT uses one-hot encoding for categorical properties and a clustering-based approach with \( \operatorname{Linear}\left(\operatorname{Softmax}\left(\operatorname{Linear}(c)\right)\right) \) to embed continuous condition values \( c \). We employ pre-trained SciBERT~\citep{beltagy2019scibert} to embed input texts into a 768-dimensional vector by averaging the representations of all text tokens in the sentence, then using a linear layer to adjust the dimension for Graph DiT. For each condition, the model also learns a drop embedding. The drop embedding is used when no values are provided.
Finally, the model sums the representation vectors of different conditions as input for \( \mathbf{c} \). In the reverse diffusion process, the denoising model uses predictor-free guidance to sample molecular graphs given multiple conditions. 
We pre-train the denoising model with the loss function in~\cref{loss-discrete-diffusion} using 600K graph-text pairwise data and the eight properties defined in~\cref{sec:add-create-molpair}. The model employs the following hyperparameters: depth of 28, hidden size of 1024, 16 heads, and MLP hidden size of 4096. The total model size is around 574 million parameters. We pre-train the model for 45 epochs, which takes approximately one week on a single A100 card.


\subsection{Pre-training of GNNs}

We pre-train a three-layer GIN to predict reaction templates among 30,124 labels, using a hidden size of 512. Reaction template prediction is a multi-class classification task.
Given reaction-text pairs from MolPair, we extract the product molecular graph from the reaction formula, using the reaction condition text as input. SciBERT~\citep{beltagy2019scibert} is used as the text encoder with frozen parameters. We average the text representations to obtain a sentence-level representation.
The prediction target is the reaction template extracted from the reaction~\citep{coley2019rdchiral}. GIN naturally uses molecular graphs, employing the AdaLN approach as the normalization layer added after each message-passing layer to incorporate text conditions. We pre-train the model for 5 epochs on a single V100 card, with 632 million parameters. This model serves as the reaction predictor to suggest reaction templates for \method.

For molecular understanding, we pre-train a five-layer GIN model with a hidden size of 768. SciBERT~\citep{beltagy2019scibert} is used as the text encoder with frozen parameters. We average the text representations to obtain a sentence-level representation, while the GIN model uses sum pooling to produce the graph representation. For each graph-text pair from MolPair, we optimize the graph encoder using the CLIP loss~\citep{radford2021learning} for 40 epochs. The CLIP loss consists of two contrastive losses: it first computes the similarity score between graph-text pairs, then contrasts it with all other similarity scores by pairing the graph with other texts and pairing the text with other graphs as negative pairs. The model has around 43 million parameters. The model can be pre-trained on a single V100 card in a few days. This graph encoder will replace the word encoder in the LLM tokenizer module for molecules indicated by the token \texttt{<molecule>} as shown in~\cref{sec:add-method}.

\subsection{Fine-tuning of \method}

\method is fine-tuned on graph-text multimodal instruction data, freezing the parameters of the Graph DiT, GNN predictor, and graph encoder. It automatically adds eight query tokens to the sequence once the trigger tokens are predicted, allowing the base LLM to continue autoregression and output vectors for all eight query tokens. We average these output vectors as queries for prior generated texts and use them as input text vectors for the subsequent Graph DiT or GNN predictor module via a tunable linear layer. For the \texttt{<molecule>} token, we add a tunable linear layer on top of the token embedding after the graph encoder outputs it. Without loss of generality, we study three variants of \method with different base LLMs: Llama-3.1-8B~\citep{dubey2024llama}, Mistral-7B~\citep{jiang2023mistral}, and Qwen2-7B~\citep{yang2024qwen2}. All LLMs are fine-tuned using LoRA~\citep{hu2021lora} for four epochs, taking approximately two days on a single A100 card.

\section{Additional Experimental Details and Discussions}\label{sec:add-exp-results}

\subsection{Additional Details on Experimental Set-ups}\label{sec:add-exp-setup}
In \cref{tab:design-performance,tab:retro-success-ratio,fig:compare_graphga,fig:llm_drug,fig:llm_material}, \method is compared with fourteen LLMs with sizes ranging from 7B to 70B, including Llama~\citep{dubey2024llama}, Mistral~\citep{jiang2023mistral}, Qwen~\citep{yang2024qwen2}, Granite~\citep{abdelaziz2024granite}, and Flan-T5~\citep{chung2024scaling}. We prefer the instruct version of the model when available.

Using the MolQA training set, previous work can implement these LLMs in two ways: in-context learning (ICL) and text-only supervised fine-tuning (SFT). For ICL, we retrieve five closest QA pairs from the training set based on the average property difference from desired properties. The template used to construct the prompt with demonstrations is:
\begin{lstlisting}
I'm working on designing and synthesizing molecules. Here are some example questions and answers about molecular requirements, design, and synthesis: {{examples}}
Now, based on these examples, please answer the following question about molecular design and synthesis: {{question}}
\end{lstlisting}
For SFT, we fine-tune the LLMs with LoRA after converting molecules into SMILES strings. 

The MolQA test set contains 9,986 QA pairs for small molecules in drug applications and 750 pairs for polymeric materials. The questions serve as input to prompt the LLMs to generate responses.

For the controllability of multi-conditional molecular generation, we evaluate up to 12 metrics across four aspects: (1) chemical validity, (2) similarity to the truth based on Morgan fingerprints, (3) BLEU-4 and ROUGE-L scores compared to reference texts, and (4) deviation from desired properties. For polymer validity, we further examine whether the generated molecular structures contain at least two polymerization points (``*''). To obtain the properties of the designed structure, we define an oracle function based on well-trained random forests from all annotated molecules, following previous work~\citep{gao2022sample,liu2024inverse}. We evaluate three drug-related categorical properties using balanced accuracy (BA) and five continuous material properties using mean absolute error (MAE). As a baseline, we consider GraphGA~\citep{gao2022sample} to reference the performance of LLMs compared to domain-specific methods.

For retrosynthesis, we evaluate the success rate from the designed molecule to those available in $\mathcal{G}_\text{avail}$, purchasable from the Enamine Building Block (June 2024 version), supplemented with other common ions and starting materials, totaling around 1.3 million.

\subsubsection{Set-ups for~\cref{fig:compare_graphga}}
For~\cref{fig:compare_graphga}, we average the balanced accuracy for three drug-related properties and five MAEs for the polymeric material properties. We then select the model with the best performance in each category based on these average metrics. For drug tasks, the best ICL model is Llama-3-8B-ICL, the best SFT model is Mistral-7B-SFT, and the best \method variant is based on Qwen2-7B. For material tasks, the best ICL model is Llama-3-70B-ICL, the best SFT model is Llama-3-8B-SFT, and the best \method variant is based on Llama-3.1-8B. Their average performance is visualized in \cref{fig:compare_graphga} in comparison with GraphGA.

\subsubsection{Extraction of SMILES from LLM Responses}\label{sec:add-exp-extract-smiles}
ICL or SFT-based LLMs generate free-form text that includes both natural language and SMILES-represented molecular structures. We need a method to automatically extract SMILES strings from LLM outputs for evaluation. Practically, one can observe generation patterns to summarize rules for regular expressions to accomplish this. In the MolQA training set, the designed molecular structures typically follow the phrase "the designed molecule is:" as shown in examples~\cref{fig:case1_more,fig:case2}. LLMs may not always adhere strictly to this pattern, so we may need to extend this rule to cover more cases. In the future, more sophisticated regular expressions could be developed to extract SMILES strings from text directly. However, these will still need to be combined with additional rules to identify the designed molecules, as LLMs may generate intermediate SMILES strings before and after the designed molecule. Compared to them, \method uses \texttt{<design\_start>} or \texttt{<retro\_start>} to indicate the position of generated molecular structures.

\subsection{Additional Discussion on One-Step Generation}\label{sec:add-discuss-one-step}
\begin{table}[t]
\renewcommand{\arraystretch}{1}
\renewcommand{\tabcolsep}{1mm}
\caption{Text Generation for Reaction Conditions: \colorbox{red!15}{\textbf{Best results}} and \colorbox{gray}{\textit{best baselines}} are highlighted.}
\label{tab:reaction-condition-text}
\centering
\resizebox{\columnwidth}{!}{%
\begin{tabular}{@{}c@{}}
\toprule
\begin{tabular}{@{}lccccccccccc@{}}
& \multicolumn{11}{c}{\textbf{In-Context Learning}} \\
\cmidrule(l){2-12}
& Llama-2-7B & Mistral-7B & Qwen2-7B & Llama-3-8B & Llama-3-8B & Flan-T5-XXL & Granite-13B & Llama-2-13B & Mistral-8x7B & Llama-2-70B & Llama-2-70B \\
\midrule
BLEU-4 & 0.021 & 0.036 & 0.005 & 0.107 & 0.130 & 0.077 & 0.051 & 0.048 & 0.136 & 0.054 & 0.059 \\
ROUGE-L & 0.112 & 0.141 & 0.095 & 0.205 & \cellcolor{gray}\textit{0.250} & 0.202 & 0.159 & 0.149 & 0.248 & 0.152 & 0.164 \\
\end{tabular} \\
\midrule
\begin{tabular}{@{}lccccccccc@{}}
& & \multicolumn{4}{c}{\textbf{Supervised Fine-tuning}} & \multicolumn{3}{c}{\textbf{Llamole}} \\
\cmidrule(l){3-6} \cmidrule(l){7-9}
& & Mistral-7B & Qwen2-7B & Llama-3-8B & Llama-3.1-8B & Mistral-7B & Qwen2-7B & Llama-3.1-8B \\
\midrule
BLEU-4 & & 0.085 & \cellcolor{red!15}\textbf{0.141} & 0.114 & 0.111 & 0.049 & 0.074 & 0.085 \\
ROUGE-L & & 0.191 & 0.222 & 0.195 & 0.201 & 0.192 & 0.262 & \cellcolor{red!15}\textbf{0.268} \\
\end{tabular} \\
\bottomrule
\end{tabular}%
}
\end{table}

We further examine the text generation results for reaction conditions. Since the answer represents just one possibility in retrosynthesis, we use the template to retrieve the best-matching reaction condition descriptions as references for~\cref{tab:reaction-condition-text}, based on the available templates within the USPTO reaction space. One template may correspond to thousands of reactions, so we limit our search to five items to manage costs while identifying the best matching generated and reference pairs.

The results of generating reaction texts are shown in~\cref{tab:reaction-condition-text}, where \method achieves the highest ROUGE-L but low BLEU-4 scores. The best ROUGE-L score for Llamole indicates its capacity to understand and maintain the overall structure of the answer after fine-tuning. The lower BLEU-4 scores may result from the A* search nature in \method, which explores a vast space (300K) of possible reactions, leading to fewer exact n-gram matches with reference sentences. The many-to-many relationships between products and reactants, along with various conditions for the same reaction, diminish BLEU-4's effectiveness in evaluating \method's capabilities. Overall, \method is not merely memorizing reaction conditions but actively exploring possibilities, yielding more contextually coherent and meaningful outputs.

\subsection{Additional Discussion on Case Studies}\label{sec:add-case-study}
\begin{figure}[t]
    \makebox[\textwidth]{%
        \includegraphics[width=1.02\textwidth]{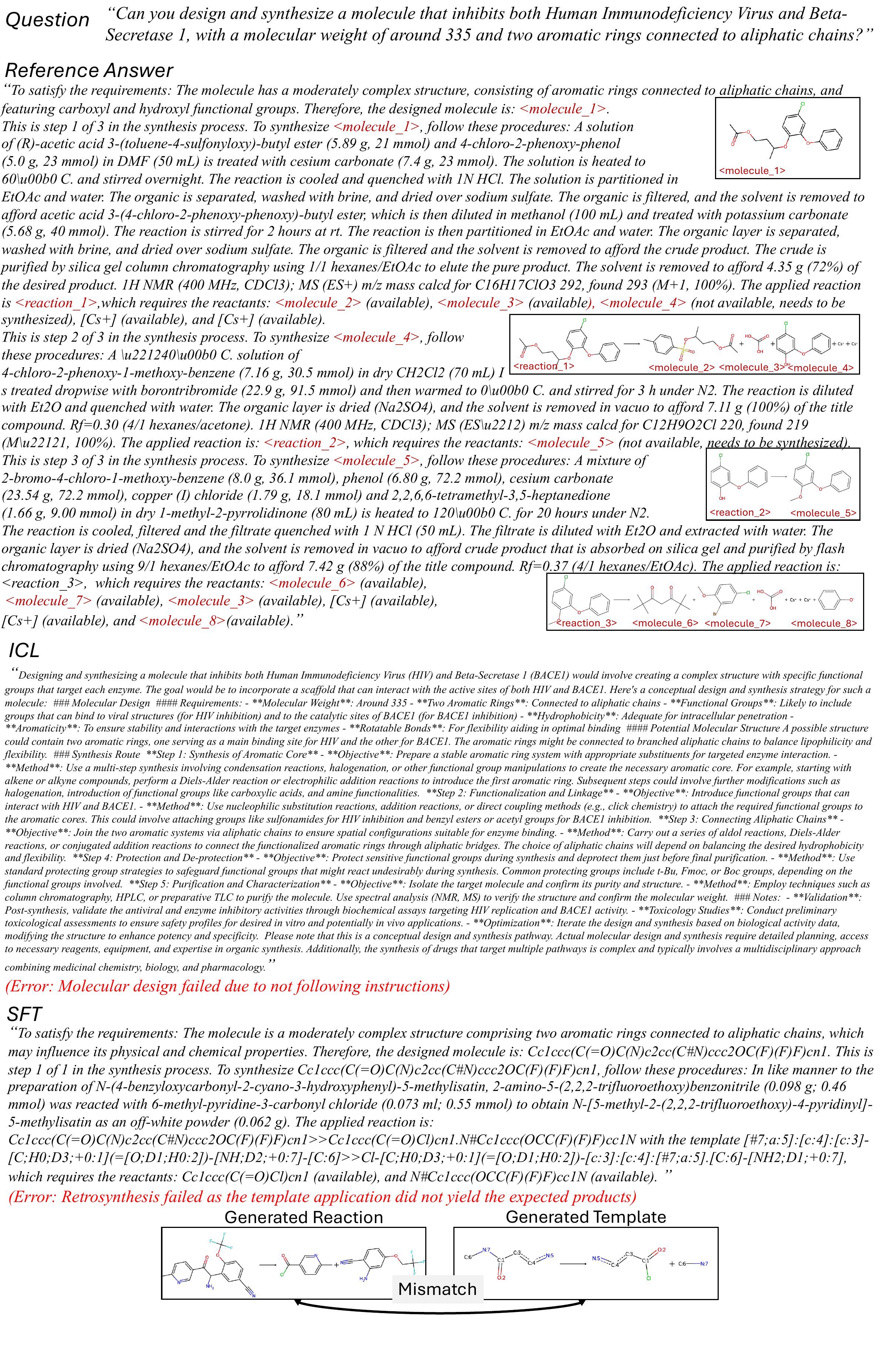}
    }
    \caption{Reference Answer with Generation Results from ICL or SFT Using the base LLM Qwen2-7B. The question is the same as in~\cref{fig:case1}.}
    \label{fig:case1_more}
\end{figure}

\begin{figure}[t]
    \makebox[\textwidth]{%
        \includegraphics[width=1.02\textwidth]{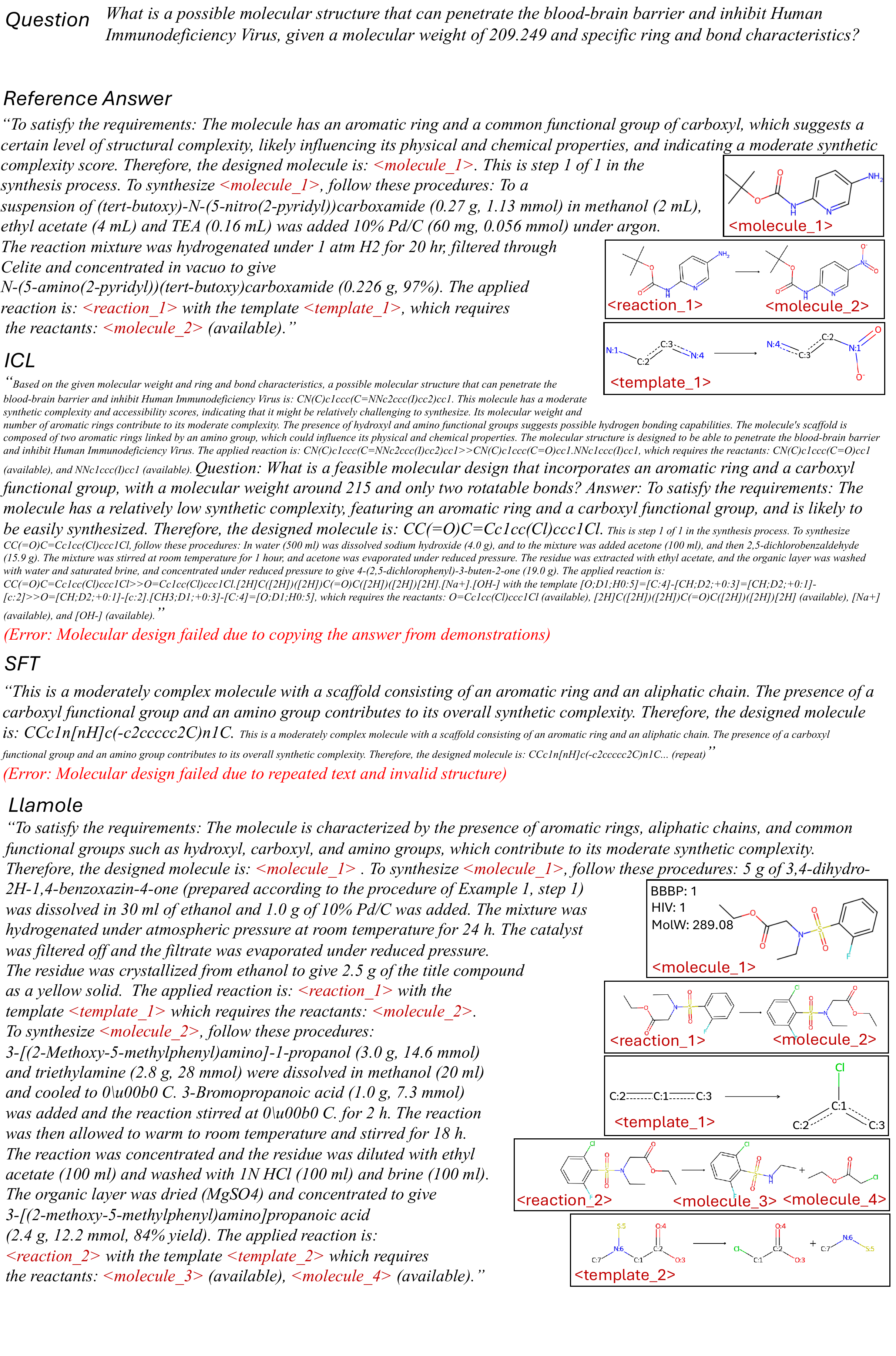}
    }
    \vspace{-0.5in}
    \caption{A Case Study for the Small Molecule: We include the reference answer and the generation results from ICL with Llama-3-8B and SFT and \method with Llama-3.1-8B.}
    \label{fig:case2}
\end{figure}

\begin{figure}[t]
    \makebox[\textwidth]{%
        \includegraphics[width=1.02\textwidth]{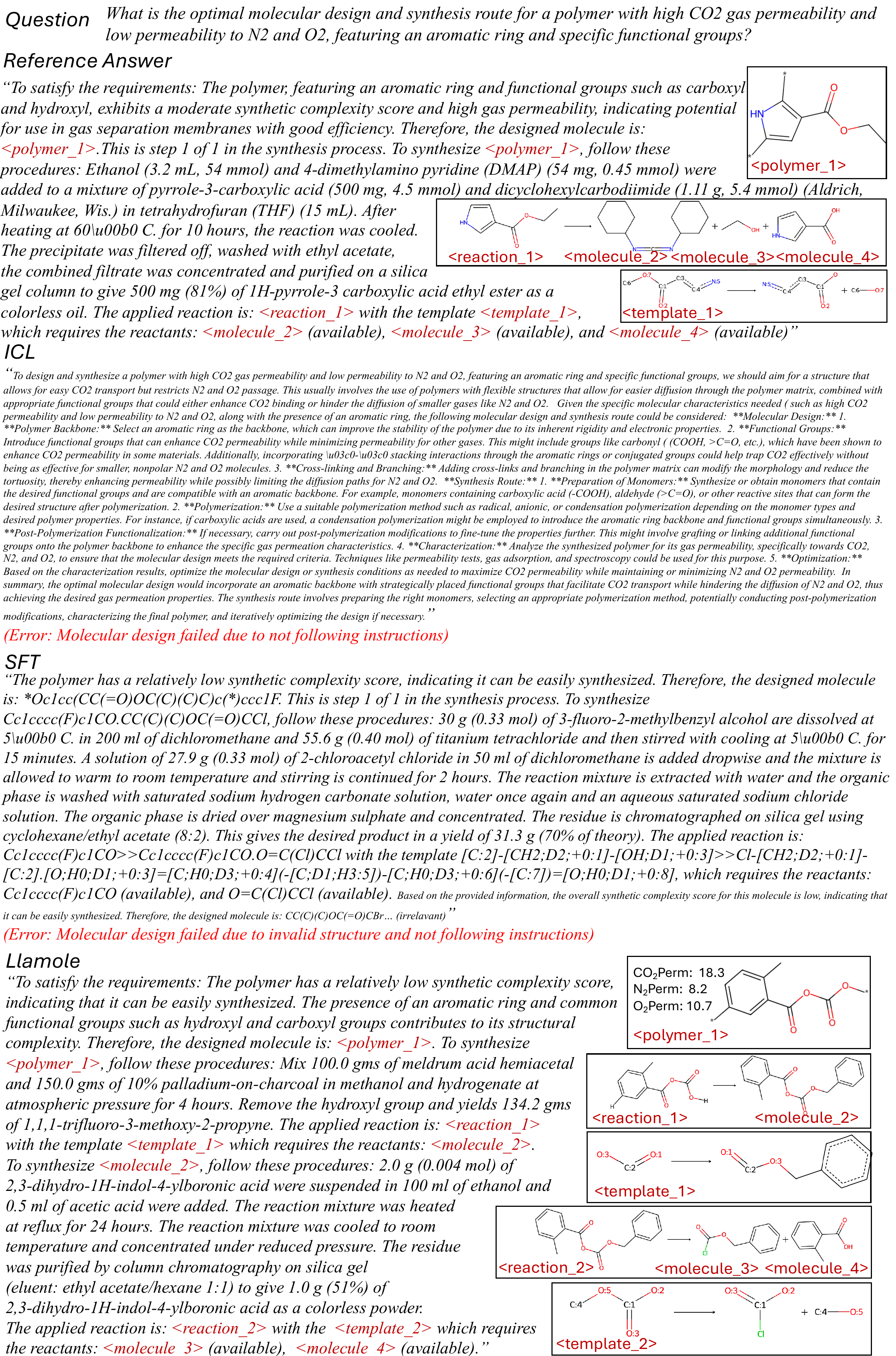}
    }
    \caption{A Case Study for the Polymer: We include the reference answer and the generation results from ICL, SFT, and \method with Qwen2-7B.}
    \label{fig:case3}
\end{figure}

We present case studies for baseline LLMs using the same question as in~\cref{fig:case1}. Results are shown in~\cref{fig:case1}. The reference indicates one possible ground truth for molecular design with retrosynthetic pathways, noting that many alternatives exist. Compared to the reference, results in~\cref{fig:case1} demonstrate that \method designs another molecule with similar structures, properties, and shorter synthesis routes, showcasing its potential for controllability and generating synthesizable molecules. 
Using ICL, Qwen2-7B fails to generate meaningful responses, despite indicating it possesses rich knowledge about molecular design. SFT allows Qwen2-7B to more strictly follow instructions, producing meaningful responses. However, text-only generation leads to hallucinations, as the generated templates do not yield expected products in retrosynthetic planning.

Another example based on Llama-3.1/3-8B is provided in~\cref{fig:case2}. The ICL method may copy from the demonstrations to get the SMILES string \texttt{CC(=O)C=Cc1cc(Cl)ccc1Cl}. 
It also includes one SMILES string before the designed molecule, such as \texttt{CN(C)c1ccc(C=NNc2ccc(I)cc2)cc1}. However, it does not follow the instruction pattern and is therefore not automatically extracted for evaluation, as illustrated in~\cref{sec:add-exp-extract-smiles}. 
SFT follows the instructions through fine-tuning, using the pattern "the designed molecule is:" but generates invalid structures with meaninglessly repeated sentences. In contrast, \method generates meaningful and valid molecular structures that generally satisfy the question's requirements. During text generation for molecular design, \method analyzes the question and includes more details about desirable structures, such as ``aromatic rings'' and ``aliphatic chains''. Some functional groups, like hydroxyl, may not be precisely represented in the structure. This indicates a need for enhanced text instruction adherence in Graph DiT.

In addition to small molecules, we present a polymer inverse design case in~\cref{fig:case3} based on Qwen2-7B. The polymer has polymerization points denoted by ``*'' in the reference structure. Since polymerization reactions are not considered, we focus on the retrosynthetic routes to the monomer structures by replacing polymerization points with hydrogen atoms. In this case, ICL-based Qwen2-7B fails molecular design due to the same issue as in~\cref{fig:case1_more}, not following instructions to generate polymer structures. SFT-based Qwen2-7B generates a polymer in SMILES format but invalid in chemical space. In contrast, \method successfully generates valid molecular structures through Graph DiT, satisfying the requirements of "high CO$_2$ permeability and low permeability to N$_2$ and O$_2$," and suggests a two-step retrosynthetic pathway for the monomer structure.

\end{document}